\documentclass[preprint,12pt,authoryear]{elsarticle}
\usepackage{amssymb}
\usepackage{multirow}
\usepackage[]{algorithm2e}
\usepackage{textcomp, gensymb}
\usepackage{float}
\usepackage{amsmath}
\usepackage[graphicx]{realboxes}
\usepackage{subcaption}
\usepackage{longtable} 
\usepackage{booktabs}
\usepackage{rotating}
\usepackage{natbib}
\usepackage{numprint}
\journal{Reliability Engineering \&\ System Safety}

\begin{document}

\begin{frontmatter}

\title{Epistemic and Aleatoric Uncertainty Quantification for Crack Detection using a Bayesian Boundary Aware Convolutional Network}

\affiliation[inst1]{organization={School for Engineering of Matter, Transport and Energy},
addressline={Arizona State University}, 
city={Tempe},
postcode={85287}, 
state={AZ},
country={USA}}

\author[inst1]{Rahul Rathnakumar}
\author[inst1]{Yutian Pang}
\author[inst1]{Yongming Liu\corref{mycorrespondingauthor}}
\cortext[mycorrespondingauthor]{Corresponding author.}
\ead{Yongming.liu@asu.edu}

\begin{abstract}

Accurately detecting crack boundaries is crucial for reliability assessment and risk management of structures and materials, such as structural health monitoring, diagnostics, prognostics, and maintenance scheduling. Uncertainty quantification of crack detection is challenging due to various stochastic factors, such as measurement noises, signal processing, and model simplifications. A machine learning-based approach is proposed to quantify both epistemic and aleatoric uncertainties concurrently. We introduce a Bayesian Boundary-Aware Convolutional Network (B-BACN) that emphasizes uncertainty-aware boundary refinement to generate precise and reliable crack boundary detections. The proposed method employs a multi-task learning approach, where we use Monte Carlo Dropout to learn the epistemic uncertainty and a Gaussian sampling function to predict each sample's aleatoric uncertainty. Moreover, we include a boundary refinement loss to B-BACN to enhance the determination of defect boundaries. The proposed method is demonstrated with benchmark experimental results and compared with several existing methods. The experimental results illustrate the effectiveness of our proposed approach in uncertainty-aware crack boundary detection, minimizing misclassification rate, and improving model calibration capabilities.

\end{abstract}

% Research highlights - SHORT
\begin{highlights}
    \item We propose a Bayesian CNN with boundary refinement for crack segmentation.
    \item Uncertainty decomposition into epistemic and aleatoric sources improves insights.
    \item Epistemic uncertainty can indicate distribution shift and insufficient training data.
    \item Uncertainty-aware crack detection improves performance compared to the FCN baseline.
\end{highlights}


\end{frontmatter}
\section{Introduction}
\label{sec:Introduction}
Cracks are a major source of failures in structural and industrial components. There are several reasons why cracks might form in these structures. For instance, roads and building cracks occur over time due to exposure to environmental factors such as humidity and rapid temperature changes. They also occur due to overload, such as heavy traffic in the case of roads. Phenomena such as corrosion can also lead to crack initiation and growth. These cracks, if undetected, can lead to catastrophic failure. Detecting cracks early is crucial to prevent such failures, improve maintenance protocols, reduce costs, and extend the lifespan of infrastructure.
\\
Vision-based inspection offers a safe, efficient and rapidly scalable solution for crack detection. Other signal-processing modalities that are popular for crack detection include millimeter-wave imaging \citep{bivalkar2022development} and ultrasonic inspection \citep{sun2023nonlinear, lee2022demonstration}. The primary benefit of vision-based inspection for crack detection is its versatility in various applications. It can detect various types of cracks depending on the sensor parameters, from hairline to large size, on a wide range of materials including composite, metal, ceramic, and plastic. Moreover, it is not affected by factors such as temperature and vibration. The primary drivers of improvements to industrial inspection technology using vision-based approaches are higher computational power, miniaturized, commercially available sensors, and rapid advancements in machine learning techniques. These factors have opened up new possibilities for continuous condition monitoring of civil infrastructure at scale. In addition, autonomous inspection techniques provide an opportunity to remove subjectivity that can result from manual inspectors. While deep learning-based techniques are getting widely adopted for various types of industrial and infrastructure inspection tasks, for our experiments, we narrow the focus down to working on crack detection in structural and infrastructural systems, such as crack detection in concrete surfaces and road pavements.
\\
Accurately detecting cracks within an image is a complex and challenging problem. Convolutional neural networks (CNNs) have been used to recognize and classify cracks, but early deep learning-based models focused solely on the classification problem and ignore underlying feature information such as crack boundaries. Detecting accurate crack boundaries is important for downstream condition monitoring and maintenance scheduling. Traditional vision-based fault detection methods use hand-crafted features that limit generalization capacity. To address this, deep learning has been used to develop more flexible and accurate models for fault detections. Given that fine-grained crack detection provides us with highly resolved morphology in an end-to-end fashion, the question of whether the detection is to be trusted is important.This is an interesting problem because the adoption of deep learning approaches can be accelerated if we can quantify the confidence of the prediction produced by a neural network and address the problem of detecting distributional shifts that occur in practice after deployment. The consequences of poor detection results in a situation where the final risk assessment may be inaccurate, resulting in either a higher frequency of catastrophic events or maintenance cost overruns. Therefore, the first goal of this paper is to produce uncertainty estimates along with predictions for detection. To do this, we make use of a sampling-based Bayesian Deep Learning approach to decompose the sources of uncertainty in prediction using the formulation proposed by \citep{kendall2017uncertainties}.  Experiments to benchmark the validity of the proposed uncertainty quantification approach are done to ensure that the characteristics needed for detection are met. The next goal is to formulate the detection problem using a multi-task loss that learns a distributional loss using the log-likelihood term, refines the boundaries of the crack, and align it to the ground truth using a boundary loss term. This formulation borrows from \cite{DBLP:journals/corr/abs-2102-02696} to compute a boundary loss that learns to better align predicted and ground truth boundaries. To demonstrate the effectiveness of the method, we conduct experiments on two commonly used crack segmentation datasets and report results on model performance, calibration, and uncertainty for both within and out of distribution samples. 

To summarize, the contributions of this paper are as follows:
\begin{itemize}
    \item We propose a Bayesian Boundary Aware Convolutional Network (B-BACN) for crack segmentation that can predict aleatoric uncertainty, provide a sampling-based epistemic uncertainty, and refined boundary using the active boundary loss.
    \item We provide detailed empirical evaluations of within and out-of-distribution cases in order to analyze the effect that additional training samples have on uncertainty, predictive performance and model calibration.
    \item We use the proposed method to analyze the improvements that boundary losses can bring to improve the accuracy of the crack morphology, and we use uncertainty as a tool for analyzing model calibration. 
    \item We argue for the use of uncertainty and model calibration as an important  performance index when assessing and comparing computer vision models in industrial settings, where  small dataset sizes and gradual distributional shift after deployment are common.
\end{itemize}
    
\section{Related Work}
\label{sec:Related Work}
\subsection{Crack Detection Techniques}
A lot of early studies that focus on the crack detection problem used methods that led to a series of improvements in detection capabilities which saw extensive feature engineering and signal processing efforts \citep{Woods_1989, Kirschke_1992, Mao-de_2007, Ayenu-Prah_2008}. The availability of more imaging data and the explosion of deep learning led to bounding box approaches for detecting cracks, which were mostly inspired by the YOLO \citep{Redmon_2016}  and R-CNN \citep{Girshick_2015, Ren_2015}. These works exploited pre-trained backbones from these networks and fine-tuned it on crack datasets \citep{deng2021imaging, Mao_2020, Hacıefendioğlu_2021, Kato_2022}. While these approaches effectively exploited the availability of higher compute power and pre-trained weights for effective fine-tuning, they did not provide end-to-end pixel-wise prediction for cracks. Fine-grained crack detection results are crucial for any detailed morphological analysis of the structures of interest, so a bounding-box approach to this problem is not sufficient. To address this, semantic segmentation approaches were proposed and evaluated on benchmark datasets, alleviating the need for hand-crafted feature engineering and classical signal processing. The Fully Convolutional Network has been an established technique for object detection and segmentation in medical imaging, introduced by \citet{long2015fully}. This approach saw rapid and extensive adoption in the industrial  inspection field, with improvements to the segmentation accuracy by introducing widely used concepts such as feature pyramid hierarchies \citep{Yang_2020} and hierarchical feature learning \citep{zou2018deepcrack, Cheng_2021}, loosely tied to the seminal works by \citet{He_2016, Ronneberger_2015}.
\subsection{Defect Boundary Reproduction and Detection}
Improving the reproduction and detection of boundaries of objects is challenging and relevant to the crack segmentation problem, as outlined in Section ~\ref{sec:Introduction}. Over the years, multiple different approaches have been proposed, including early traditional methods such as active contours \citep{Chan_2001}. Early work in boundary refinement also included applying post-processing methods such as the Conditional Random Field (CRF) based method proposed by \citep{krahenbuhl2011efficient} and image filtering to refine crudely localized crack boundaries and then classify them using classical ML models, as seen in \cite{Shi2016}. Since crack pixels are similar to edges, early works that used filters were inspired by edge detectors. However, these methods did not have semantic knowledge of the crack pixels and their relation to the background. This meant that a lot of post-processing and advanced filter design is required to remove false positives. Later works began to utilize approaches that defined boundary refinement blocks \cite{chen2020supervised}. \citep{guo2021barnet} adapts the original image gradient with the coarse crack detection result and refines it to precise crack boundaries.  
\subsection{Uncertainty Quantification for Reliable Condition Monitoring}
%%% ADD THE DETAILS FOR RELIABILITY PERSPECTIVE HERE ^^^ %%%
Estimating model uncertainty remains a significant topic of interest in a wide array of applications such as condition monitoring \citep{Moradi_2022, Seites-Rundlett_2021}, fault diagnostics \citep{Zhou_2022}, and remaining useful life estimates \citep{Zhu_2022} A lot of the works in reliability engineering focuses on utilizing predictive models to perform system-wide reliability assessments. \citep{Moradi_2022} uses a Bayesian Network to model a system-wide network and analyze its component risks using a data-driven framework. After the deployment of a predictive model, it is vulnerable to distribution shifts over time. \citep{Zhou_2022} proposed a Bayesian Deep Learning technique to structural component health using 1-D signals, considering potential OOD samples. Uncertainty Quantification (UQ) is used as a way to classify these samples. Uncertainties have also been to query for informative samples for active learning techniques \citep{yang2016active, wang2018uncertainty}. \citep{Zhu_2022} uses this technique to predict battery degradation. Uncertainty quantification under distributional shifts for fault diagnostics has previously been studied by \citep{HAN2022108648} using an ensemble of neural networks, where the authors use a thresholding approach to determine whether the uncertainty associated with a prediction makes it more likely for the sample to have originated from outside the training distribution. \citet{Sajedi_2020} incorporated uncertainty into their structural health monitoring model by using Monte-Carlo Dropout sampling and prediction quality classification and showed that variance of softmax and entropy correlate with misclassification rate. Well calibrated uncertainties are key for condition monitoring, especially when using neural networks, which are notorious for providing overconfident estimates. Improving calibration has been studied by \citep{https://doi.org/10.48550/arxiv.1706.04599}, where the authors propose temperature scaling to tune the outputs of a neural network. Obtaining calibrated predictions for regression tasks using neural networks has been studied by \citet{Tohme_2022}, where the authors propose a novel loss function formulation. 
\subsection{Bayesian Approaches for Defect Detection}
While novel neural network architectures have improved model performance for crack detection, uncertainty estimates for crack detection has not been  studied to the same extent. Deep learning techniques do not lend themselves to analytical Bayesian inference and require approximate inference techniques, which have seen a lot of development in computational statistics literature \citep{blei2017variational, zhang2018advances}. Among Bayesian Deep Learning methods, there have been a multitude of approximate evaluation methods proposed, including Variational Bayes-By-Backprop \citep{blundell2015weight}, Monte Carlo Dropout \citep{gal2016dropout}, Spectral-Normalized Gaussian Process (SNGP)\citep{lakshminarayanan2020simple}. These approaches have been adopted in a variety of applications \citep{pang2022bayesian, pang2021data, lee2017deep, FAN2023109088} to estimate epistemic uncertainty. However, the quantification and decomposition of both epistemic and aleatoric uncertainty in the defect characterization context has been missing, with more attention being paid in the field to collecting domain-specific datasets and improving predictive performance. Therefore, our contribution decomposes epistemic and aleatoric uncertainties for crack characterization, and uses the  uncertainty to assess model performance and calibration. 
\\
The study of uncertainty decomposition has recently been discussed by \citet{McFarland_2020} in the context of risk assessment. \citet{McFarland_2020} have focused on decomposing the epistemic and aleatoric components for problems such as cracking defect failure probability using classical Bayesian techniques. In order to extend the literature on uncertainty for defect characterization, our work aims to use the dropout approximation to Bayesian inference in neural networks to extract epistemic uncertainty \citep{srivastava2014dropout, gal2016dropout} and a negative log likelihood formulation to predict aleatoric uncertainty. A recent work similar to ours on the uncertainty quantification front is \citep{Pyle_null}, where they provide uncertainty analyses for crack characterization using ultrasonic sensors for in-line inspection. \citep{Pyle_null} studies uncertainty on a specialized ultrasonic crack characterization dataset, and comments on uncertainty analyses using various quantification methods for out-of-distribution (OOD) cases. However, our work focuses on using uncertainty to analyze improvements to model performance and calibration on OOD and within-distribution data using a boundary-refinement objective while training the model. 
\\
In summary, the crack detection model in our work combines uncertainty decomposition and boundary refinement techniques to produce detection results that result in two main benefits: the ability to accurately segment crack morphology, and the ability to separately compute epistemic and aleatoric variances. The uncertainty is used to analyze prediction confidence and whether the model is well-trained for the test set being considered. This work also analyzes the effect that additional training samples have on uncertainty, and provides detailed empirical evaluations of within and out-of-distribution cases. 
\section{Methodology}
\label{sec:Methodology}
% INSERT network architecture details
\begin{sidewaysfigure}
    \includegraphics[width=1.0\textwidth]{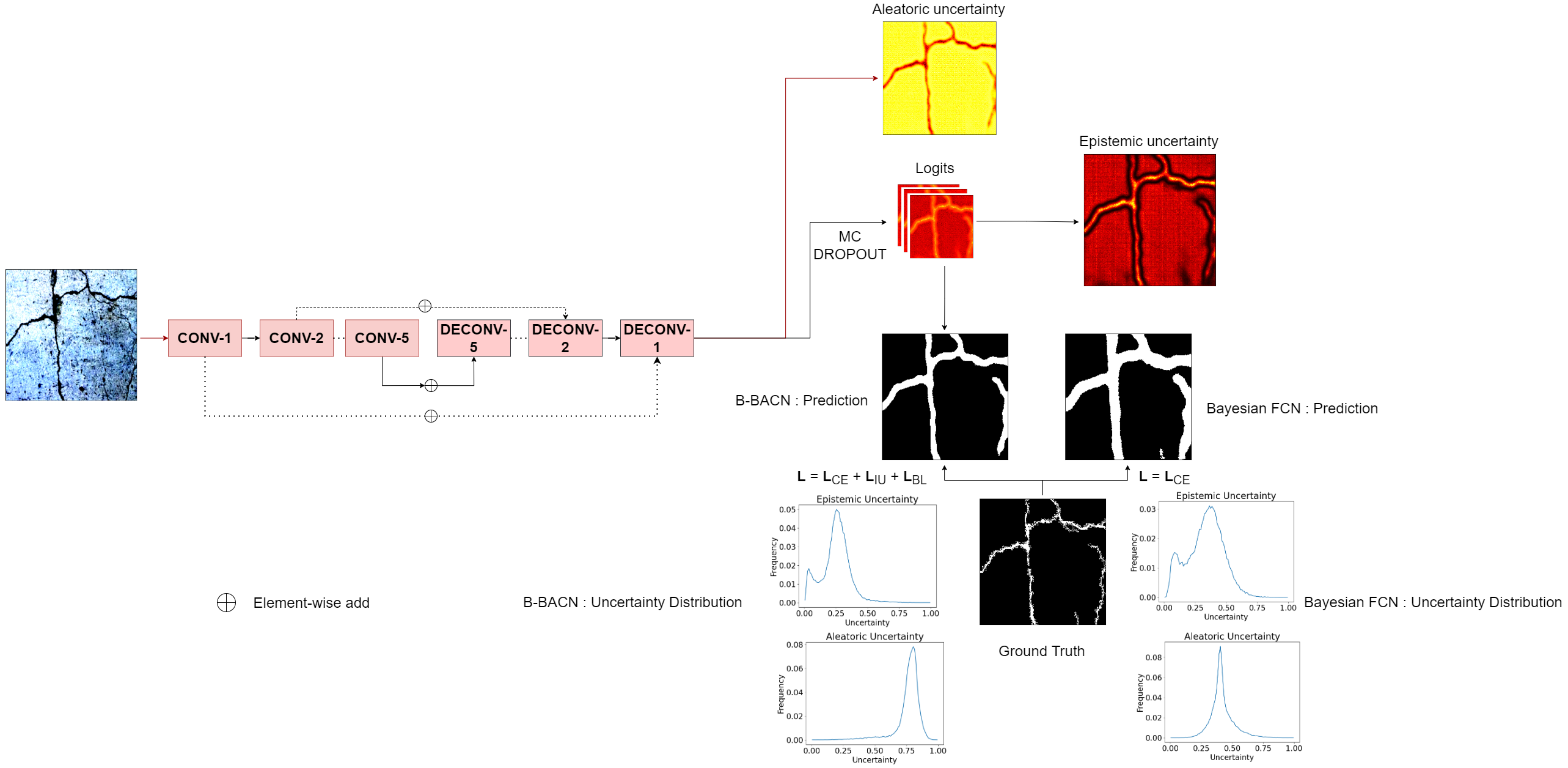}
    \caption{Overview of the proposed network architecture showing the network structure, input, predictions and uncertainty. The use of the proposed approach allows the model to predict tighter crack boundaries and epistemic uncertainties. 
}
    \label{fig:netarchitecture}
\end{sidewaysfigure}
\begin{figure}[htp]
  \centering
  \begin{minipage}[b]{0.45\linewidth}
    \centering
    \includegraphics[width=\linewidth]{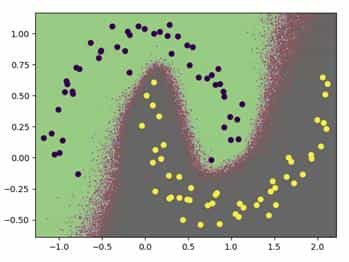}
    \caption{Classification with uncertainty in the two moons dataset - Variation of uncertainty and entropy with sample size}
    \label{fig:twomoons}
  \end{minipage}
  \hspace{0.05\linewidth}
  \begin{minipage}[b]{0.45\linewidth}
    \centering
    \includegraphics[width=\linewidth]{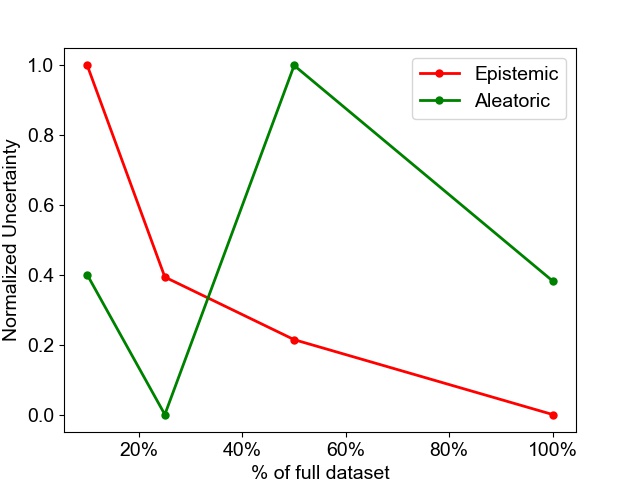}
    \caption{Classification with uncertainty in the two moons dataset - Variation of uncertainty with sample size.}
    \label{fig:classificationToy_bars}
  \end{minipage}
\end{figure}

The overall architecture is based on the fully-convolutional encoder-decoder network \citep{long2015fully}, with a Resnet-50 backbone, as shown in Figure~\ref{fig:netarchitecture}. The encoder network consists of 5 blocks, with pre-trained weights from the ImageNet dataset that are retrained for each training run. The encoder blocks provide a hierarchy of features at multiple scales, with earlier layers extracting fine-grained morphological information, and later layers extracting coarse-grained category and location information. The resulting layers are then passed into the decoder, which consists of transposed convolutions such that the feature outputs mirror the corresponding size of the next encoder layer. In addition to having information flow sequentially through each layer in the network,the skip-connections across the encoder-decoder structure combines the representations obtained across multiple scales. This approach has been successfully used in the past to help with boosting gradient flow, improving the vanishing gradient problem.
\\
Structural prognostics benefit immensely from being able to model predictive uncertainty at multiple stages of analysis. Point estimates provided by deep neural networks may lead to overconfident predictions that can often be wrong. Ideally, we seek expressive models that help extend the capabilities of the model for downstream tasks and have a way to explain its predictions to the end user. We use a Bayesian Deep Learning approach to accomplish this task. 
\\
Consider the general setup for Bayesian inference:
\begin{equation}
    p(y^* \vert x^*,X,Y)= \int p(y^* \vert f^* )p(f^* \vert x^*,X,Y)df^*
\label{eq: equation_bayes}
\end{equation}
In Equation \ref{eq: equation_bayes}, a prior distribution is defined over the space of functions p(f) where f is a sample from the distribution of functions that could have generated the data  \(D = (X,Y) \). The likelihood function is \(p(Y \vert f,X)\). The posterior can be written using the Bayes rule from the prior p(f) and the likelihood function. For an unseen data point \(x^*\), the posterior predictive can be computed by integrating over all possible functions \(f^*\) to yield the conditional probability for prediction \(y^*\). The problem with Equation \ref{eq: equation_bayes} arises from the inability of neural networks to produce any meaningful uncertainty estimate as part of the prediction. While it is easy to specify conjugate distributions for simpler statistical models, it is far more challenging to obtain closed form posterior distributions for neural networks. This is because of the large dimension of the weight matrix, which leads to intractable integrals.
\\
Approximate Variational Inference (VI) methods estimate the Bayesian posterior by minimizing a metric between an approximate distribution and the true posterior \citep{blei2017variational}:
\begin{align}
    \theta^* = argmin_{\theta} KL[q(w|\theta) || P(w)] - E_{q(w|\theta)}[log P(D|w)]
    \label{eq:VI_EQN}
\end{align}
We consider these methods in our study, with a major focus on MC Dropout (MCD) with a predictive aleatoric variance output learned during training. The other methods we compare this against are Concrete Dropout (CD) \citep{YarinConcrete2017} and Bayes By Backprop (BBB) \citep{blundell2015weight}. The MCD approximation is shown to be a special case of the generic VI approximation in \citet{gal2016dropout}. 
\\
%Write down the equation forms of MCD, GD, CD, and BBB
The BBB approach uses a sampling approach to compute the KL divergence in Equation~\ref{eq:VI_EQN}, using the approximating distribution \(q_\theta (w|D)\) to draw samples \(w^{(i)}\) from and optimize the following objective:
 \begin{equation}
     F(D,\theta) \approx \sum_{i = 1}^{n} log q (w^{(i)}|\theta) - log P(w^{(i)}) - log P(D|w^{(i)})
 \end{equation}
The MCD approach is an approximate VI method with a specific choice of the approximating distribution \(q_\theta(w|D)\). In this case, \(q_\theta\) is a distribution over the weight matrices whose columns are randomly set to zero, which is exactly what dropout does:
\begin{align}
    W_i &= M_i.diag([z_{i,j}]_{j=1}^{K_i}) \\
    \nonumber
    z_{i,j} &\sim Bernoulli(p_i),   i = 1, ... , L j = 1, ..., K_{i-1}
\end{align}
MCD model inference minimizes the KL-Divergence between \(q_\theta(w|D)\) and a deep Gaussian process. Details on this derivation are given in the Appendix of \citet{gal2016dropout}. After inference, we evaluate the uncertainty by averaging \( N \) forward passes during prediction. This is equivalent to drawing \( N \) samples from the set of parameters defined in the model posterior and evaluating a function using each of those samples. The sample mean of the predictive distribution approaches the population means at large sample sizes, with a spread that is dictated by the form of the function resulting from each dropout sampling iteration. The question here is whether the uncertainty is purely derived from the uncertainty about the weights. The analysis of the variance by \cite{brach2020single} provides some insight into this problem: 
\begin{equation}
    \begin{split}
        E_i^D &= E_i(1-p^*) \\
        V_i^D &= V_i p^*(1-p^*) + V_i (1-p^*)^2 + E_i^2 p^*(1-p^*)
    \end{split}
    \label{eq: equation 13}
\end{equation}
Equation \ref{eq: equation 13} shows that the expectation and variance of the output \(E_i^D\) and \(V_i^D\) only depend on the expectation \(E_i\) and variance \(V_i\) from the previous layer and the applied dropout ratio p.  This shows that the only contribution to the variance in MC Dropout is from the dropout layer. Therefore, this component of uncertainty is driven by the uncertainty in the weights. 
\\
The dropout probability \(p_i\) is a hyperparameter that requires expensive grid search to obtain a good epistemic uncertainty.  We implement the algorithm used by \citep{YarinConcrete2017}, which optimizes the dropout probability p using an entropy regularization term. 
% For a dense layer, this is given by:
% \begin{align}
%     KL(q_\theta(w)||P(w))&\alpha \frac{l^2(1-p)}{2}||M||^2 - KH(p) \\
%     \nonumber
%     H(p) &= -plogp - (1-p)log(1-p)
% \end{align}
% , where \(M\) is the mean weight matrix, \(l\) is a length scale parameter and \(H\) is the entropy regularization term.
\\
We now discuss how the dropout objective function operates using an example of a model trained using a squared error loss. In this case, dropout regularization is equivalent in expectation to the ridge regression model. The input feature X \( \epsilon R^{N \text{ x } D} \) is to be transformed to a target Y \( \epsilon R^{N \text{ x } 1} \), using weights w \( \epsilon R^{D  \text{ x } 1} \). Applying dropout implies multiplying the feature element-wise with a dropout mask matrix Z $\sim$ Bernoulli(p). Casting this with dropout applied to the feature in the mean-squared error loss gives: 
\begin{equation}
    L(w) = argmin_w E_{(Z \sim Bern(p))} ||y - (Z \odot X) w||^2
    \label{eq: regressionLossWithDropout}
\end{equation}
% Each element in R can either be 0 or 1, with an expectation E(R) = p and a variance var(R) = p(1-p). Since X is not random, the expectation \( E(R \odot X) = pX \) and variance \( var(R \odot X) = p(1-p)\Gamma^2 \), where \( \Gamma = diag(X^TX)^{1/2} \).
Computing the expectation of the squared error loss gives us the following:
\begin{equation}
    L(w) = ||y - X \tilde{w}||^2 + \frac{1-p}{p}||\Gamma \tilde{w}||^2,
    \tilde{w} = p.w
\end{equation}
Directly comparing the loss for the dropout with the L2 regularized loss shows that it has stronger restrictions on its Lagrange multiplier term. Concretely, \( \lambda = \frac{1-p}{p} \) for dropout, where \( p \) is the dropout probability hyperparameter, compared to the looser restriction on the multiplier for the ridge loss, \(\lambda >= 0 \). 
\\
Classical Bayesian linear regression formulations as seen in \cite{bishop2006pattern} assume a constant, known value for the aleatoric uncertainty term. However, this assumption does not hold for real-world data, as some training samples can have higher variance than others. To remedy this, our model learns a component of the variance as a function of the data, referred to as the heteroscedastic aleatoric uncertainty - the component of uncertainty that cannot be explained away with more data. The epistemic component of the uncertainty is calculated using Monte Carlo (MC) dropout during inference. In the context of applying uncertainty quantification to detection, epistemic uncertainty plays a more important role than aleatoric uncertainty, unlike in natural image situations with a lot of data, where the aleatoric uncertainty matters more. Sources of variability such as occlusion can be considered “input-dependent” and are captured by the heteroscedastic aleatoric uncertainty term. On the other hand, the industrial imaging domain does not always have a lot of samples to train a model on, which requires epistemic uncertainty to reflect this. The approach used by \citep{kendall2017uncertainties} distinguishes the epistemic and aleatoric uncertainty by deriving the loss function from the likelihood formulation. We demonstrate this approach first on a toy classification problem.
\\
Epistemic and aleatoric uncertainty are related to the decision boundary in classification. We verify the approach on the two moons dataset, visualized in Figure~\ref{fig:twomoons}. The two moons dataset is a non-linearly separable dataset with 2 inputs. We use the two-moons dataset because this is a low-dimensional toy problem that can be demonstrated with a large set of training samples to clarify the role of epistemic and aleatoric uncertainty. Since the model has 2 categories, the network outputs 4 values, 2 for the prediction logits and 2 for each of the variances. The variance terms are the input-dependent uncertainties. For a classifier:
\begin{equation}
\begin{split}
    y_i | x_i \sim \mathsf{Bern}(\psi(w^T x_i)) \\
\end{split}
\label{eq: equation 14}
\end{equation}
Assuming softmax activation $\psi$, the optimal weights can be derived using Maximum Likelihood Estimation (MLE):
\begin{subequations}
\begin{align}
    w &= \mathsf{argmax}_w p(y_i, x_i|w)  \label{subeq: equation 15a} \\
    w &= \mathsf{argmax}_w p(y_i | x_i, w) p(x_i | w) \label{subeq: equation 15b}  \\
    w &= \mathsf{argmax}_w p(y_i | x_i, w) \label{subeq: equation 15c} 
\end{align}
\label{eq: equation 15}
\end{subequations}
Equations (\ref{subeq: equation 15a})-(\ref{subeq: equation 15c}) lead to a stochastic negative log-likelihood loss by first setting up a Gaussian distribution to sample from. The mean of this distribution is taken as the prediction \(f(x_i)^w \). The variance term comes from the learned aleatoric variance output from the network. We then sample \(t\) times from this distribution and then average its log-softmax over the sampling dimension. To avoid underflow and overflow issues, this is implemented by first transforming the \(y_{i,t} \) using the log sum exp trick and then averaging over the sampling dimension. The final form of the loss is implemented directly using the \textit{Negative Log-Likelihood (NLL)} loss, after obtaining the samples as described, and comparing the prediction against the ground truth. The loss is stochastic because it depends on Monte Carlo draws from the variance term.
\begin{equation}
\begin{split}
    y_i|w &\sim N(f(x_i)^w, \sigma_i^{w^2})  \\
    y_{i,t} &= f(x_i)^w + \epsilon_t , \epsilon_t \sim N(0,  \sigma_i^{w^2}) \\
    \mathcal{L}_{MLE} &= NLL(y_{i},y_{gt})
\end{split}
\end{equation}
During the model evaluation, the procedure to compute uncertainty is by doing Monte Carlo sampling using dropout to obtain predictions. The epistemic uncertainty is computed as the variance of these predictions. The aleatoric uncertainty is the learned component of the prediction and is directly obtained from the formulation. Figure~\ref{fig:classificationToy_bars} shows that the epistemic uncertainty reduces with more training data while the aleatoric uncertainty remains oscillatory and independent of this trend. Figure~\ref{fig:twomoons} shows a demonstrative example of what the decision boundary looks like when using MC-Dropout. The areas with more data have a lower uncertainty in the decision boundary, indicated by the lower amount of variance in classification. At the extreme ends of the decision boundary, we notice a widening of the noise profile, indicating a higher uncertainty for the discriminative model to assign a fixed label to a sample near the boundary.
The next component is the active boundary loss (ABL) that further refines predictions, proposed by \citep{DBLP:journals/corr/abs-2102-02696}. Since the loss is model-agnostic, we adopt it for our problem and provide a brief description of it here. This loss improves the alignment between predicted boundaries (PDBs) and ground-truth boundaries (GTBs) during training by moving PDBs toward the closest GTBs. The ABL loss is differentiable and dynamic and it focuses on the relationship between PDB and GTB pixels. It can be used in combination with other loss terms such as cross-entropy loss and Lovasz-softmax loss \citep{https://doi.org/10.48550/arxiv.1705.08790} to improve the boundary details in image segmentation. The method can be useful in preserving the boundaries of thin objects in an image. The ABL continuously monitors changes in the PDBs to determine the plausible moving directions. The method is divided into two phases: First, for each pixel i on the PDBs, the next candidate boundary pixel j closest to the GTBs is determined. Second, the KL divergence is used to encourage the increase in KL divergence between the class probability distribution of i and j. Meanwhile, this process reduces the KL divergence between i and the rest of its neighboring pixels. In this way, the PDBs can be gradually pushed toward the GTBs. Unfortunately, conflicts might occur and the performance of the ABL can degrade, so the authors use gradient flow control to reduce the conflicts. Further details of this process are elaborated in \citep{DBLP:journals/corr/abs-2102-02696}. We also use the Jaccard loss, given by the Lovasz-softmax function \citep{ https://doi.org/10.48550/arxiv.1705.08790}, to refine the regional shape of the detected crack by directly optimizing for higher Intersection-over-Union (IoU) score. 
\\
Additionally, we perform temperature scaling on the logits to  calibrate the predicted probability distributions of the network, following \cite{https://doi.org/10.48550/arxiv.1706.04599}, where the temperature is a learnable hyperparameter ($T$). We use temperature scaling to calibrate the confidence of the model, by scaling the logits of the mean prediction head. We do not scale the aleatoric uncertainty predictions. The temperature scaling procedure is summarized as follows:
\begin{enumerate}
\item Obtain the pre-softmax outputs logits of the trained neural network for the validation set.
\item Iterate over a range of temperature scalar values and select the one that minimizes the NLL loss over the validation set using the L-BFGS optimizer.
\item Divide the logits by the optimal temperature scalar.
\end{enumerate} 
In summary, the overall proposed loss is the combination of the derived heteroscedastic classification loss, the IoU loss and the ABL loss:
\begin{equation}
    L = L_{MLE} + L_{ABL} + L_{IOU}
    \label{eq:loss_simple_add}
\end{equation}
\section{Experimental Setup}
\label{sec:ExperimentalSetup}
\subsection{Implementation Details}
The experiments are performed on the CrackForest dataset by \cite{Shi2016} and DeepCrack dataset \citep{LIU2019139}. The CrackForest dataset is representative of a variety of cracks encountered in urban roads, and has been used as a benchmark for multiple studies such as \cite{Shi2016}, \cite{fan2018automatic}, \cite{yang2019feature}, \cite{zou2018deepcrack}. It consists of 100 images, with 18 left out for the test set. The DeepCrack dataset is a larger dataset consisting of crack images at multiple scales and background textures. It consists of 443 training set images and 78 test set images. 
% [Training details - Network architecture, optimizer details, batch size, epochs and learning rate scheduling]
The network architecture used for the main experiments is the Resnet50 backbone encoder with a transposed convolutional decoder, as explained in Section \ref{sec:Methodology}, and the models are trained using PyTorch. The training code has the ability to run in both single and multi GPU configurations, with our training using a single NVIDIA 1080Ti GPU that has 11 GB of VRAM. The encoder uses a set of pre-trained weights from the ImageNet dataset, which is then fully retrained using the specific dataset for the experiment. We employ the use of the SGD with momentum optimizer, with a momentum value of 0.9 and weight decay of \(10^{-5}\), along with a step-decay learning rate scheduler across our experiments. For the initial learning rate, we start at 0.01, which decays every 50 epochs by a factor of 0.8. During training, we also implement early stopping based on the validation loss history over the past 50 epochs. The early stop condition used for our experiments is an average validation loss reduction of less than 0.001. For the experiments across training samples, a dropout probability of 0.5 was used. We extract the epistemic uncertainty during model evaluation using 25 MC runs through the network. 
% Revision 1: Loss function weighting ablation study
\\
As part of our experiments, we also include a study on how the weighting of the three losses influences performance. Specifically, we consider three strategies:
\begin{enumerate}
\item Simple linear superposition (Baseline): Here, all three losses are directly added up, as seen in Equation~\ref{eq:loss_simple_add}
\item Ramp-up boundary loss weighting (Ramp): In this case, we ramp-up the boundary loss terms using a sigmoid ramp-up function. The ramp-up ends after 100 epochs, determined empirically based on the loss convergence of the cross-entropy loss:
\begin{equation}
    L = L_{MLE} + \phi(t)(L_{ABL} + L_{IOU}), 
\end{equation}
where \(\phi\) is the sigmoid function, and t is the epoch number.
\item Coefficient of Variations Weighting (CoV) \citep{groenendijk2020multiloss}: The method is founded on the Coefficient of Variation, which is the ratio of the standard deviation to the mean and shows the extent of variability of the observed losses in relation to their mean:
\begin{subequations}
\begin{align}
    \label{eq:cov_a} L &= \alpha_{1t} L_{MLE} + \alpha_{2t} L_{ABL} + \alpha_{3t} L_{IOU} \\
    \label{eq:cov_b} \alpha_{it} &= \frac{c_{lit}}{z_t} = \frac{1}{z_t}\frac{\sigma_{lit}}{\mu_{lit}} \\
    \label{eq:cov_c} l_{it} &= \frac{L_t}{\mu_{L_{t-1}}}
\end{align}
\end{subequations}
Equation~\ref{eq:cov_a} represents the weighted total loss function, composed of three terms weighted by \(\alpha_{1t}, \alpha_{2t}, \alpha_{3t} \). Equation~\ref{eq:cov_b} defines the weighting term \(\alpha_{it}\) for each component, defined as the coefficient of variation of the loss ratio \(l_{it}\). The loss ratio \(l_{it}\), found to be a robust training strategy by \citet{groenendijk2020multiloss}, measures the ratio of the current loss value to the mean of the loss history \(\mu_{L_{t-1}}\), as seen in Equation~\ref{eq:cov_c}.
\end{enumerate}
\begin{table}[ht]
\centering
% \small
\caption{Architecture details of the segmentation network}
\begin{tabular}{llllll}
\toprule
\textbf{Layer type} & \textbf{Num. filters} & \textbf{Filter size} & \textbf{Stride} & \textbf{Padding} & \textbf{Activation} \\ \midrule
ResNet-50 Backbone & - & - & - & - & - \\
Conv2d & 1024 & 3 & 1 & 1 & ReLU \\
BatchNorm2d & 1024 & - & - & - & - \\
ConvTranspose2d & 512 & 3 & 2 & 1 & ReLU \\
BatchNorm2d & 512 & - & - & - & - \\
ConvTranspose2d & 256 & 3 & 2 & 1 & ReLU \\
BatchNorm2d & 256 & - & - & - & - \\
ConvTranspose2d & 128 & 3 & 2 & 1 & ReLU \\
BatchNorm2d & 128 & - & - & - & - \\
Dropout & - & - & - & - & - \\
Conv2d & - & 1 & - & - & - \\ \bottomrule
\end{tabular}
\label{tab: layerDetails}
\end{table}
\begin{table}[h]
\centering
\caption{Training details and hyperparameters}
\begin{tabular}{ |p{7cm}|p{8cm}| }
\hline
\textbf{Hyperparameter} & \textbf{Value} \\
\hline
Optimizer & SGD with momentum \\
\hline
Momentum & 0.9 \\
\hline
Initial learning rate & 0.01 \\
\hline
Learning rate scheduler & Step Decay Learning Rate \\
\hline
Learning rate epoch decay interval & 50 \\
\hline
Decay factor & 0.8 \\
\hline
Early stop criterion & validation loss \\
\hline
Early stop - Loss history & 50 epochs \\
\hline
Early stop - Average loss reduction & 0.001 \\
\hline
\end{tabular}
\label{tab: trainConfig}
\end{table}

To quantify model performance, we make use of the macro F-1 score to report the mean F-1 score, which is an unweighted average of the class-wise F-1 scores.
\subsection{Evaluation Metrics}
To quantify model performance, we make use of the macro F-1 score to report the mean F-1 score, which is an unweighted average of the class-wise F-1 scores. The F-1 score is defined  as the harmonic mean of precision and recall and typically used to measure how well the segmentation captures crack morphology. 
\\
The equation for the F-1 score is:
\begin{equation}
Precision =  \frac{True~Positive}{True~Positive + False~Positive}
\end{equation}
\begin{equation}
Recall = \frac{True~Positive}{True~Positive + False~Negative}
\end{equation}
\begin{equation}
    F_1 = 2 \cdot \frac{Precision \cdot Recall}{Precision + Recall}
    \label{eq:F1Score}
\end{equation}

We also compute the entropy and variance of the predictions over the MC runs to assess the level of uncertainty in the model out of those samples. 
The entropy is computed per-class by averaging the Monte Carlo mean prediction over each pixel as:
\begin{equation}
H(y) = -\sum_{c=1}^{C}y_c \cdot \log y_c
\label{eq:entropy}
\end{equation}
where $C$ is the number of classes and $y_i$ is the logit output of class $i$. The epistemic variance is computed across the set of MC predictions:
\begin{equation}
Var(y) = \frac{1}{M} \sum_{i = 1}^{M} (y_i - \Bar{y})^2
\end{equation}
wher $M$ is the number of MC samples, \(y_i\) is the sampled output from the network, and \(\Bar{y}\) is the mean prediction.
\\
To quantify the model's calibration, we make use of the Expected Calibration Error (ECE) \citep{https://doi.org/10.48550/arxiv.1706.04599} \citep{naeini2015obtaining}, using 30 bins to measure the difference between the calibration score and the reliability.  The ECE is computed using Eq.~\ref{eq:ECE}, with the calibration bin accuracy \( Acc(B_m) \) and the overall accuracy of the class \(m\).
\begin{equation}    
ECE = \sum^M_{m=1} \frac{|B_m|}{n}|acc(B_m) - conf(B_m)| 
\label{eq:ECE} 
\end{equation}
\section{Results and Discussion}
\label{sec:Results}
\begin{figure}
    \centering
    \includegraphics[width=15cm]{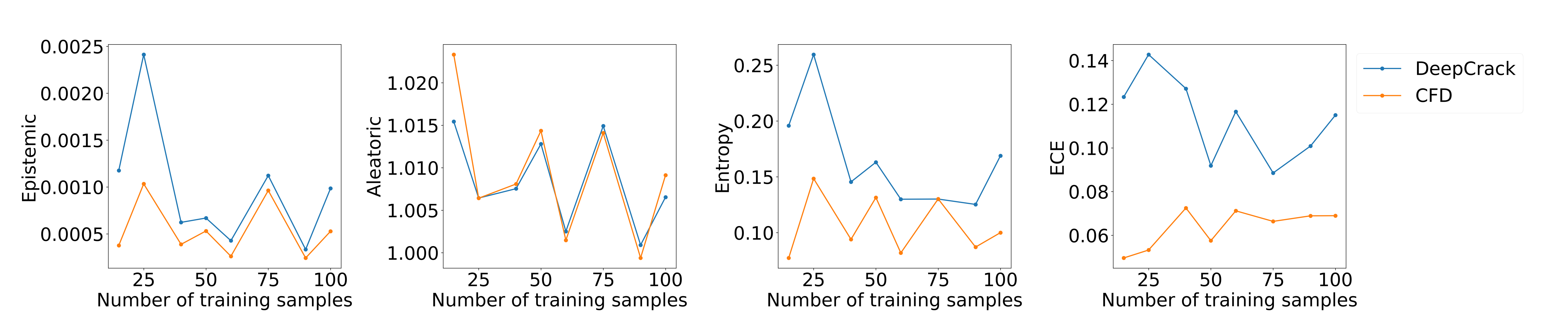}
    \caption{Evaluation of models trained on the CFD dataset across training samples on both the CFD test set and the DeepCrack test set.}
    \label{fig:trainingSamples_CFDTrain_DeepCrackCFDEval}
\end{figure}

\begin{figure}
    \centering
        \includegraphics[width=15cm]{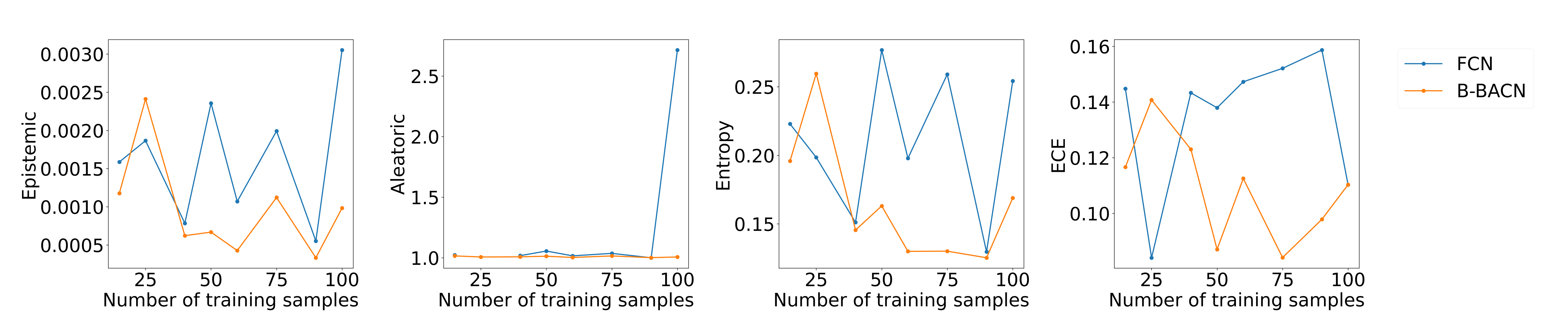}
    \caption{Evaluation of models trained on the CFD dataset across training samples on the DeepCrack test set on the Cross-Entropy trained Bayesian FCN baseline and the proposed B-BACN model.}
    \label{fig:trainingSamples_CFDTrain_DeepCrackEval_BaselineVsABLIOU}
\end{figure}

\begin{figure}[h]
    \centering
    \includegraphics[width=\linewidth]{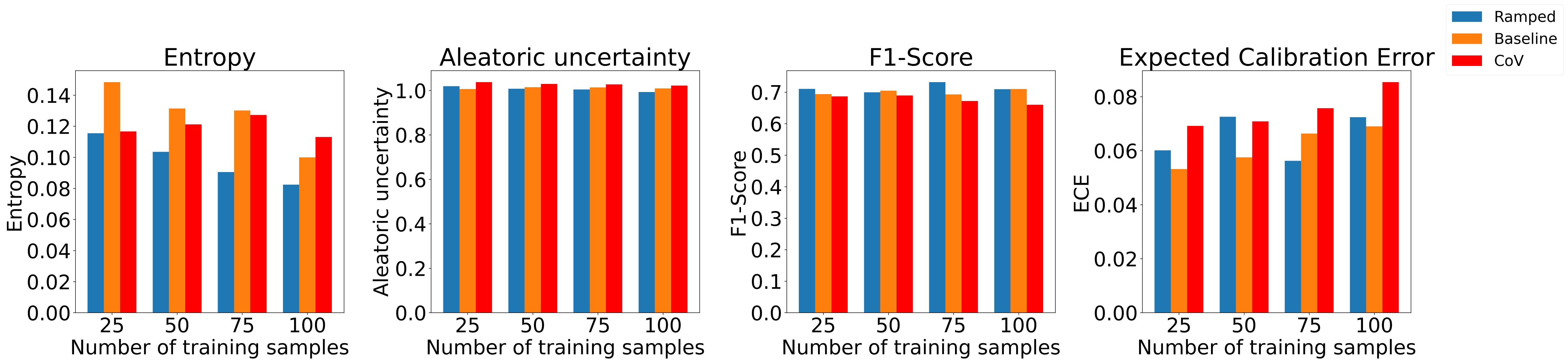}
    \caption{Model Performance Comparison on CFD Dataset with Varying Training Samples: Epistemic and Aleatoric Uncertainty, F1-Score, and ECE. }
    \label{fig:LossWeighting_ablation_CFD}
\end{figure}

In our experiments, we first show the performance of the model using the baseline FCN model, the B-BACN model and the Temperature-Scaled B-BACN model on the DeepCrack dataset and the CrackForest datasets, establishing that the proposed B-BACN model significantly improves the detection performance, while reducing uncertainty and improving the expected calibration error. The technique used for modeling uncertainty, as demonstrated in Section \ref{sec:Methodology}, aims to capture specific behaviors of uncertainty, such as detecting higher uncertainty when the test samples are farther from the training samples and having an invariant aleatoric component with respect to the training samples. Therefore, we demonstrate model performance across training samples and dropout, revealing the utility of uncertainty metrics for determining whether the model is well-trained and to what extent these trends seen on the simpler two-moons dataset hold for the segmentation dataset. We then demonstrate how the uncertainty model generalizes to unseen and out-of-distribution data samples from the CrackForest dataset. Since the DeepCrack dataset is out of the distribution of the CrackForest dataset, we expect the epistemic entropy to increase for these samples. Additionally, we show that the aleatoric component of uncertainty does not decrease despite training on the CrackForest dataset. We also present the performance of our model on a smaller CrackForest dataset, trained with noisy data, with and without the introduction of the boundary loss. We evaluate the model's performance quantitatively using metrics such as the F1 score, as well as qualitatively by using demonstrative detection results and uncertainty maps. 
\\
% What were the results of using a boundary loss over the vanilla cross entropy loss? How was this data obtained? Why is it important?
\begin{table}
 \caption{Performance comparison for models trained on DeepCrack and CrackForest dataset}
\hskip-2.0cm
\begin{tabular}{llrrrrr}
\toprule
  Dataset &                     Model &  F1-score &  Epistemic &  Entropy &  Aleatoric &     ECE \\
\midrule
DeepCrack &                       FCN (\cite{long2015fully}) &   0.70890 &    0.00032 &  0.10525 &    \textbf{1.02133} & 0.14657 \\
DeepCrack &                    B-BACN &   \textbf{0.80060} &    \textbf{0.00019} &  \textbf{0.08406} &    1.03108 & 0.10371 \\
DeepCrack & Temperature Scaled B-BACN &   0.79581 &    0.00020 &  0.13612 &    1.03223 & \textbf{0.09010} \\
\hline
      CFD &                       FCN (\cite{long2015fully}) &   0.63862 &    0.00054 &  0.13309 &    \textbf{1.03619} & 0.09943 \\
      CFD &                    B-BACN &   \textbf{0.69882} &    \textbf{0.00022} &  \textbf{0.08280} &    1.01341 & 0.07617 \\
      CFD & Temperature Scaled B-BACN &   0.69690 &    0.00025 &  0.14228 &    1.01340 & \textbf{0.05900} \\
\bottomrule
\end{tabular}
\label{tab:PerfCompare_DeepCrackCFD}
\end{table}
\begin{figure}[h]
    \centering
    \includegraphics[width = 10cm]{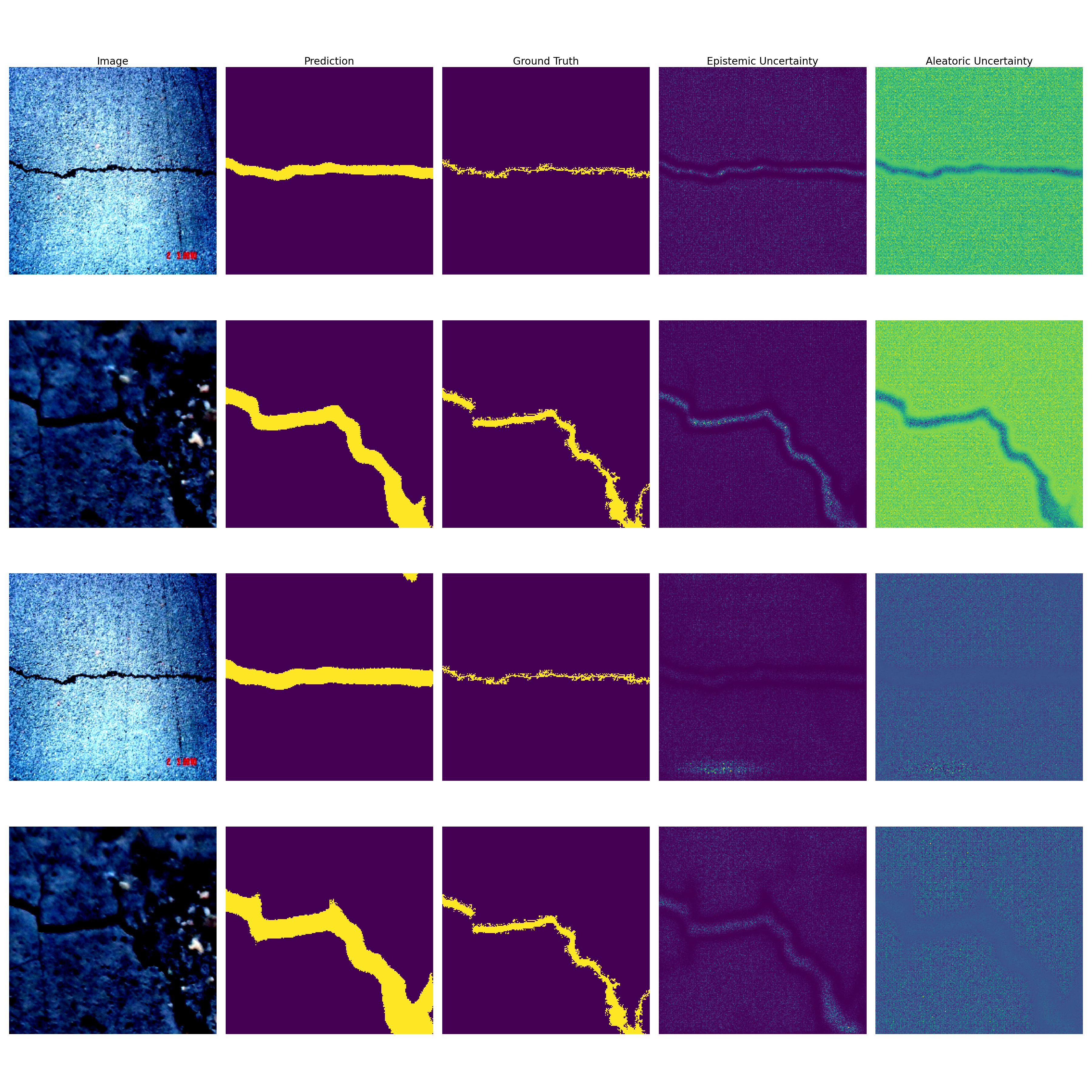}
    \caption{Demonstrative examples of detections in the CrackForest (Top-2 rows) and the DeepCrack Dataset (Bottom-2 rows). Images from Left to Right: Input, Prediction, Ground Truth, Epistemic Uncertainty, Aleatoric Uncertainty}
    \label{fig:demonstrativeResults_UncertaintyMaps_CrackDetection}
\end{figure}
The first set of results, shown in Table \ref{tab:PerfCompare_DeepCrackCFD}, compares model performance on the DeepCrack test set using the boundary loss against the MC Dropout baseline. Not only does the boundary loss exceed F1 score compared to the baseline, but also shows a drop in the variance of the prediction. The models used were trained on the DeepCrack dataset using all the available training data. These results indicate that using the boundary loss terms helps to directly operate on and refine the pixel-level predictions. 
Temperature scaling enhances the Expected Calibration Error (ECE) without altering the predictions themselves. It achieves this by scaling each logit individually, while preserving the maximum logit as the maximum value. This is why we do not see a significant change in the F-1 score or the epistemic uncertainty. Similar trends are observed for the CrackForest dataset.
\\
Figure ~\ref{fig:demonstrativeResults_UncertaintyMaps_CrackDetection} provides some demonstrative detection examples that compare these results. The predicted boundary changes were also compared between the boundary loss model and the cross entropy model, with the boundary loss model providing more accurate boundaries and lower uncertainty compared to the baseline.
\\
Evaluating the model across various training samples shows a trend similar to the one seen in the toy problem in Section \ref{sec:Methodology}. We plot these results in Figure \ref{fig:trainingSamples_CFDTrain_DeepCrackCFDEval} and Figure \ref{fig:trainingSamples_CFDTrain_DeepCrackEval_BaselineVsABLIOU}. The epistemic uncertainty showed a rather jagged trend for the baseline cross-entropy trained model, but showed a stronger decreasing trend for the B-BACN. This result was somewhat surprising as we expect there to be an approximately inverse relationship between the amount of training data and the uncertainty. Given enough training samples, more information does not meaningfully change the uncertainty. \citep{CHEN20101764} shows that the result holds true only for normally distributed variables, and that the relationship between the uncertainty and information is not straightforward in other cases. The level of uncertainty is higher on the out-of-distribution DeepCrack dataset while the aleatoric uncertainty, as expected, does not change much. The calibration error is also higher in the out-of-distribution dataset, and adding more training samples from the CrackForest dataset does not improve this. 
\begin{figure}[htb]
  \centering
  \begin{subfigure}{\textwidth}
    \includegraphics[width=\linewidth]{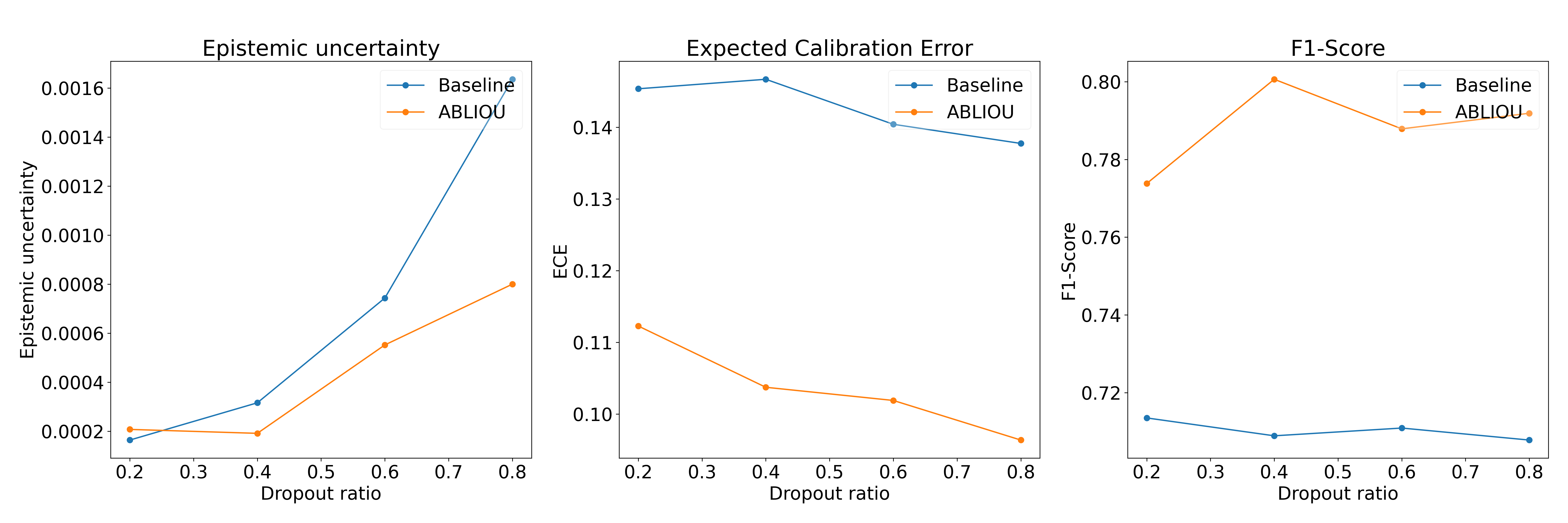}
    \caption{DeepCrack Dataset}
  \end{subfigure}
  \hfill
  \begin{subfigure}{\textwidth}
    \includegraphics[width=\linewidth]{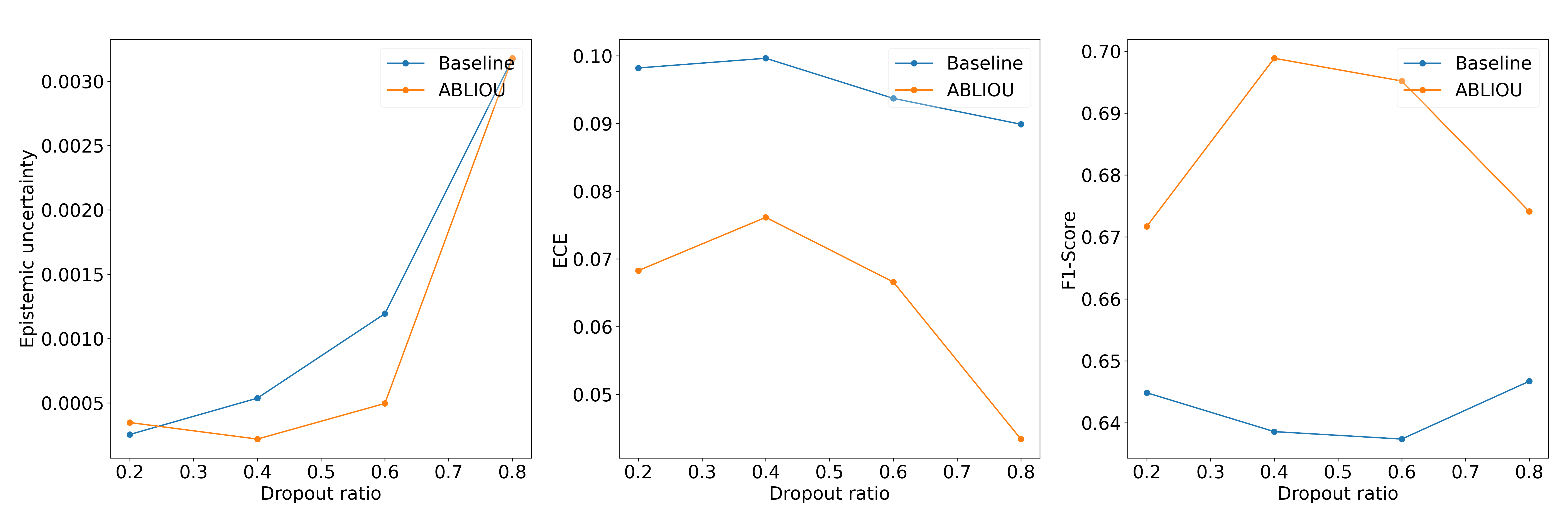}
    \caption{CrackForest Dataset}
  \end{subfigure}
  \caption{Uncertainty, F1 score and calibration error on the DeepCrack and CFD test sets for models trained on the two respective datasets across dropout ratios}
\label{fig:DropoutRatio_DeepCrackTrainDeepCrackEval}
\end{figure}
\\
Analyzing the models with various dropout ratios reveals the effect of modifying this important hyperparameter. Increasing dropout was expected to make the model less prone to overfit but also increase the prediction variance - This expectation bore out across both the baseline MC Dropout model, and the B-BACN. We also observe a consistent pattern of lower uncertainty, better calibration, and higher predictive performance across dropout ratios for the B-BACN model, as shown in Figure~ \ref{fig:DropoutRatio_DeepCrackTrainDeepCrackEval}.
%% INSERT NEW EXPERIMENT RESULTS FROM REVISION 1 %%
% loss weighting strategies % 
\\
Next, we analyze the effect of loss weighting strategies on model performance. Figure~\ref{fig:LossWeighting_ablation_CFD} demonstrates that using a simple linear superposition of losses results in a higher epistemic entropy in most cases. The differences in aleatoric uncertainty and F-1 score are not significant. However, the sigmoid ramp-up technique produces marginally better predictive performance compared to all other techniques. The higher F-1 score when trained with 75 samples compared to 100 samples using the ramp-up technique is likely an artefact. Surprisingly, the model with the least F-1 score and the highest ECE was the CoV technique across all training samples. 
\\
% Dropout layers ablation study %
\begin{figure}
    \centering
    \includegraphics[width=16cm]{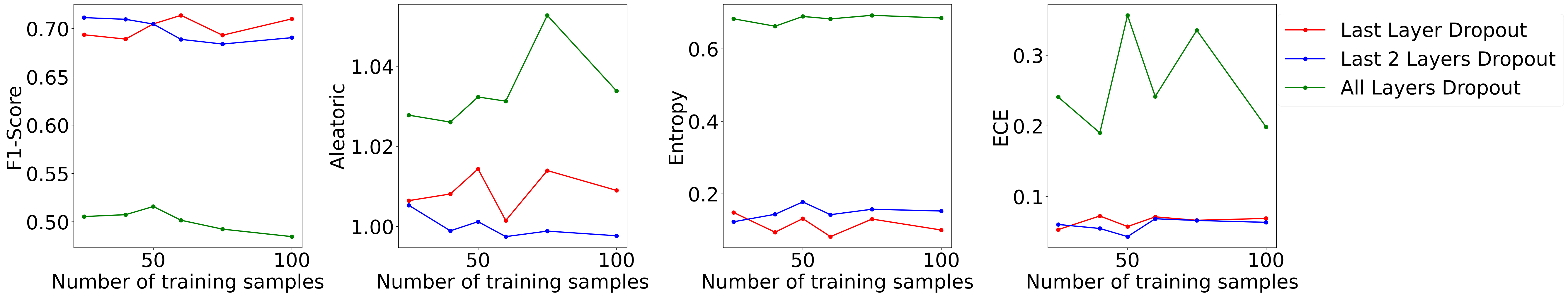}
    \caption{Effect of varying the number of dropout layers in the network on model performance and uncertainty.}
    \label{fig:DropoutHyperParameter_CFD}
\end{figure}
We then study the effect that the number of dropout layers has on model performance and uncertainty. We apply dropout on either just the final layer, the last two layers, or all the layers of the network decoder. Generally, we find that the F-1 score of the predicted observations is highest when the number of dropout layers is limited to either just the final layer, or the last two layers. Due to the strong regularization effect offered by a dropout probability of 0.5 across all decoder layers, the F-1 score falls significantly. The aleatoric uncertainty is also higher for a network that is strongly regularized by dropout. We find that the predictive entropy is significantly higher on the strongly regularized model, and the model is also less calibrated than the weakly regularized models in the other two cases. 
\begin{figure}[ht]
    \includegraphics[width=11cm]{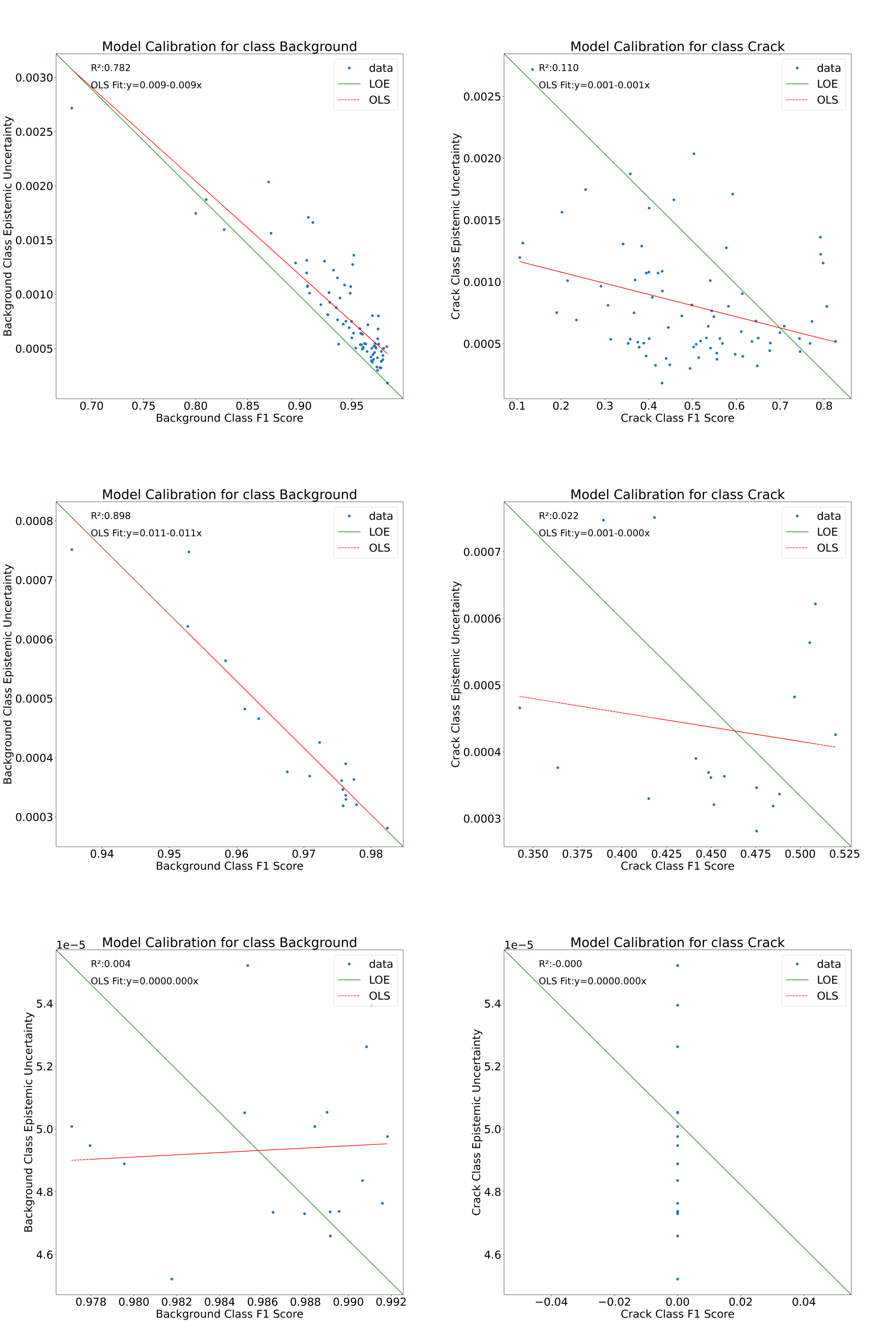}
    \caption{Model Calibration Plot for a model trained on the DeepCrack dataset : (Top row) Evaluation on the DeepCrack test set (Middle row) Evaluation on the CrackForest test set (Bottom row) Evaluation on a noisy CrackForest test set}
    \label{fig:uncertaintyCalibration}
\end{figure}
Next, we by provide a more detailed demonstration of model calibration using the model trained on the DeepCrack dataset. To do this, we first assessed the calibration on the DeepCrack test set, then evaluated the model's performance on the CrackForest test set. which contains out-of-distribution examples. Finally, we tested the model's robustness by introducing noise to the CrackForest test set to simulate a challenging scenario. One of the main objectives of predictive models is to ensure that a higher level of uncertainty is assigned to labels that deviate from the ground truth assignment, which can be verified in the validation set as the ground truth is available during the evaluation process. Figure~ \ref{fig:uncertaintyCalibration} illustrates how the class-wise model uncertainty score aligns with the prediction quality, which is quantified by the class-wise F1 score. The points should be close to the line of equality (LOE) as shown in green, or demonstrate a negative correlation to uncertainty, which indicates that there is increasing uncertainty for samples that have a poor detection score in the test set. The first row of Figure~ \ref{fig:uncertaintyCalibration} demonstrates that the DeepCrack background data is very well calibrated, but the crack class is less calibrated, with far more scatter across the predictive uncertainty. However, the calibration performance worsens with the out-of-distribution data in row 2. In row 3, however, we add Gaussian noise with a mean of 0 and a variance between 10 and 50. Introducing such high levels of noise shows the strong effect that it has on the model calibration. Model calibration methods such as the one shown in this study can therefore be used along with the expected calibration error to obtain more detailed evaluations of the image quality.
% UQ Comparison
\\
We end the results section with a comparison of the MC-Dropout (MCD) method against the Concrete Dropout (CD) and the Bayes By Backprop (BBB) methods, and also mention some limitations of the MCD method that was the focus of this study. Table~\ref{tab:PerfCompare_UQMethods} shows model performance, model uncertainty and calibration error for models trained and tested on the CFD and DeepCrack datasets. We observe that model performance does not significantly change for the CFD dataset. However, in the DeepCrack dataset, concrete dropout shows significantly higher F-1 score and significantly lower epistemic uncertainty. While the BBB method showed the best ECE metric, it slightly underperformed on both tasks in terms of F-1 score while producing a wider variance compared to the dropout-based models (underlined in Table ~\ref{tab:PerfCompare_UQMethods}). 
\begin{table}
 \caption{Performance comparison across uncertainty quantification (UQ) methods for models trained on CrackForest and DeepCrack datasets}
% \hskip-2.0cm
\begin{tabular}{llrrr}
\toprule
UQ Method &    Dataset &        F1 &  Epistemic &       ECE \\
\midrule
MCD \citep{kendall2017uncertainties} &        CFD &  \textbf{0.710104} &   0.000529 &  0.068983 \\
CD \citep{YarinConcrete2017} &        CFD &  0.700254 &   \textbf{0.000090} &  0.075523 \\
BBB \citep{blundell2015weight} &        CFD &  0.700200 &   \underline{0.005829} &  \textbf{0.059555} \\
\hline
MCD \citep{kendall2017uncertainties} &  DeepCrack &  0.761592 &   0.001005 &  0.122964 \\
CD \citep{YarinConcrete2017} &  DeepCrack &  \textbf{0.811375} &   \textbf{0.000041} &  0.099586 \\
BBB \citep{blundell2015weight} &  DeepCrack &  0.781900 &   \underline{0.005478} &  \textbf{0.070398} \\
\bottomrule
\end{tabular}
\label{tab:PerfCompare_UQMethods}
\end{table}
While the dropout-based methods outperformed the BBB method, it should be noted that MC-dropout does not provide a posterior that directly depends on the number of samples n \citep{verdoja2021notes}, such that the posterior narrows to zero at the limit of large data. MC-Dropout does provide larger variances and Concrete Dropout somewhat fixes this by driving dropout ratio \(p\) to zero in the limit of large data. Indeed, in our experiments, we initialized the dropout ratio at 0.5, which then converged at 0.1 for both the CFD and DeepCrack datasets at the largest training data size. Moreover, we also notice empirically in our work that the epistemic variance obtained does not strictly reduce towards zero with more data. This behavior is better replicated in toy datasets but does not scale well to high dimensional datasets like those used in image segmentation. Ultimately, our work adds to the evidence that MC Dropout is a method that can be implemented with ease and can be modified to provide tighter uncertainty estimates. We also show that MC Dropout is a useful technique to provide indications of dataset shift. However, we also note that it comes with its share of downsides, and future work should focus on UQ methods that can more faithfully replicate a stronger increase in uncertainty with lower training data and in regions far away from the training data regime. 
\clearpage
\section{Conclusion}
\label{sec:Conclusion}
In this paper, we have introduced the Bayesian-Boundary Aware Convolutional Network (B-BACN), which incorporates uncertainty decomposition into an epistemic and aleatoric component. Our approach shows significant performance improvements, while reducing uncertainty and improving the expected calibration error. The utilization of boundary loss functions in the B-BACN model not only refines pixel-level predictions but also contributes to reducing model size.
\\
Our analysis highlights important implications for reliability engineering, as it offers insights into the performance and calibration of crack detection models. By accurately quantifying uncertainty, our model aids in making more informed decisions regarding system reliability and maintenance. 
\\
We also compared different Bayesian Deep Learning methods, including MC-Dropout and Bayes By Backprop, and found that Concrete Dropout exhibited the lowest variance predictions and achieved superior performance on the DeepCrack dataset. These findings suggest that leveraging advanced Bayesian Deep Learning techniques can enhance the reliability of crack detection models. Furthermore, our investigation into distribution shift revealed that as test data deviated from the training distribution, model uncertainty increased and calibration decreased, highlighting the importance of accounting for distribution shifts in monitoring systems. 
\\
We saw that MC-Dropout provides lower variance predictions compared to Bayes By Backprop, but the lowest variance prediction was obtained by Concrete Dropout, which also achieved higher performance when evaluated on the DeepCrack dataset. Loss weighting strategies have also been shown to be key for optimizing learning, with the sigmoid ramp-up strategy for boundary losses providing much lower prediction uncertainties and marginally better F-1 scores than a simple linear superposition of losses. More involved loss weighting schemes such as Coefficient of Variance weighting did not show significant improvements to detection performance or model calibration. 
\\
While we saw an increase in the model uncertainty and a decrease in model calibration with out-of-distribution samples, we also obtained empirical evidence for a more complicated relationship between the epistemic uncertainty, the amount of training data present, and the aleatoric uncertainty in the data that was predicted by the model.  This observation can help guide future studies to provide more empirical evidence on this relationship and extend existing theoretical work.

\section{Acknowledgement}
\label{sec:Acknowledgement}
This work was supported by the Competitive Academic Agreement Program (CAAP) of the Pipeline and Hazardous Materials Safety Administration (PHMSA), which is a part of the US Department of Transportation. The CAAP project contract ID number is 693JK31950002CAAP.
%  TO DO



% \nocite{*}
\clearpage
\bibliographystyle{elsarticle-harv} 
\bibliography{cas-refs-RESS.bib}

\begin{thebibliography}{60}
\expandafter\ifx\csname natexlab\endcsname\relax\def\natexlab#1{#1}\fi
\providecommand{\url}[1]{\texttt{#1}}
\providecommand{\href}[2]{#2}
\providecommand{\path}[1]{#1}
\providecommand{\DOIprefix}{doi:}
\providecommand{\ArXivprefix}{arXiv:}
\providecommand{\URLprefix}{URL: }
\providecommand{\Pubmedprefix}{pmid:}
\providecommand{\doi}[1]{\href{http://dx.doi.org/#1}{\path{#1}}}
\providecommand{\Pubmed}[1]{\href{pmid:#1}{\path{#1}}}
\providecommand{\bibinfo}[2]{#2}
\ifx\xfnm\relax \def\xfnm[#1]{\unskip,\space#1}\fi
%Type = Article
\bibitem[{Ayenu-Prah et~al.(2008)Ayenu-Prah, Attoh-Okine, Attoh-Okine and
  Attoh-Okine}]{Ayenu-Prah_2008}
\bibinfo{author}{Ayenu-Prah, A.Y.}, \bibinfo{author}{Attoh-Okine, N.},
  \bibinfo{author}{Attoh-Okine, N.O.}, \bibinfo{author}{Attoh-Okine, N.O.},
  \bibinfo{year}{2008}.
\newblock \bibinfo{title}{Evaluating pavement cracks with bidimensional
  empirical mode decomposition}.
\newblock \bibinfo{journal}{EURASIP Journal on Advances in Signal Processing}
  \DOIprefix\doi{10.1155/2008/861701}.
%Type = Misc
\bibitem[{Berman et~al.(2017)Berman, Triki and
  Blaschko}]{https://doi.org/10.48550/arxiv.1705.08790}
\bibinfo{author}{Berman, M.}, \bibinfo{author}{Triki, A.R.},
  \bibinfo{author}{Blaschko, M.B.}, \bibinfo{year}{2017}.
\newblock \bibinfo{title}{The lovász-softmax loss: A tractable surrogate for
  the optimization of the intersection-over-union measure in neural networks}.
\newblock \URLprefix \url{https://arxiv.org/abs/1705.08790},
  \DOIprefix\doi{10.48550/ARXIV.1705.08790}.
%Type = Book
\bibitem[{Bishop and Nasrabadi(2006)}]{bishop2006pattern}
\bibinfo{author}{Bishop, C.M.}, \bibinfo{author}{Nasrabadi, N.M.},
  \bibinfo{year}{2006}.
\newblock \bibinfo{title}{Pattern recognition and machine learning}.
  volume~\bibinfo{volume}{4}.
\newblock \bibinfo{publisher}{Springer}.
\newblock \URLprefix \url{https://dl.acm.org/doi/10.5555/1162264},
  \DOIprefix\doi{10.5555/1162264}.
%Type = Article
\bibitem[{Bivalkar et~al.(2022)Bivalkar, Agarwal and
  Singh}]{bivalkar2022development}
\bibinfo{author}{Bivalkar, M.}, \bibinfo{author}{Agarwal, S.},
  \bibinfo{author}{Singh, D.}, \bibinfo{year}{2022}.
\newblock \bibinfo{title}{Development of an efficient approach for detection
  and measurement of crack length in ceramic tile manufacturing using
  millimeter-wave imaging}.
\newblock \bibinfo{journal}{NDT \& E International} \bibinfo{volume}{129},
  \bibinfo{pages}{102656}.
\newblock \DOIprefix\doi{https://doi.org/10.1016/j.ndteint.2022.102656}.
%Type = Article
\bibitem[{Blei et~al.(2017)Blei, Kucukelbir and
  McAuliffe}]{blei2017variational}
\bibinfo{author}{Blei, D.M.}, \bibinfo{author}{Kucukelbir, A.},
  \bibinfo{author}{McAuliffe, J.D.}, \bibinfo{year}{2017}.
\newblock \bibinfo{title}{Variational inference: A review for statisticians}.
\newblock \bibinfo{journal}{Journal of the American statistical Association}
  \bibinfo{volume}{112}, \bibinfo{pages}{859--877}.
\newblock \DOIprefix\doi{https://doi.org/10.1080/01621459.2017.1285773}.
%Type = Inproceedings
\bibitem[{Blundell et~al.(2015)Blundell, Cornebise, Kavukcuoglu and
  Wierstra}]{blundell2015weight}
\bibinfo{author}{Blundell, C.}, \bibinfo{author}{Cornebise, J.},
  \bibinfo{author}{Kavukcuoglu, K.}, \bibinfo{author}{Wierstra, D.},
  \bibinfo{year}{2015}.
\newblock \bibinfo{title}{Weight uncertainty in neural network}, in:
  \bibinfo{booktitle}{International conference on machine learning},
  \bibinfo{organization}{PMLR}. pp. \bibinfo{pages}{1613--1622}.
\newblock \DOIprefix\doi{https://doi.org/10.48550/arXiv.1505.05424}.
%Type = Article
\bibitem[{Brach et~al.(2020)Brach, Sick and D{\"u}rr}]{brach2020single}
\bibinfo{author}{Brach, K.}, \bibinfo{author}{Sick, B.},
  \bibinfo{author}{D{\"u}rr, O.}, \bibinfo{year}{2020}.
\newblock \bibinfo{title}{Single shot mc dropout approximation}.
\newblock \bibinfo{journal}{arXiv preprint} \DOIprefix\doi{arXiv:2007.03293}.
%Type = Article
\bibitem[{Chan et~al.(2001)Chan, Vese and Vese}]{Chan_2001}
\bibinfo{author}{Chan, T.F.}, \bibinfo{author}{Vese, L.A.},
  \bibinfo{author}{Vese, L.A.}, \bibinfo{year}{2001}.
\newblock \bibinfo{title}{Active contours without edges}.
\newblock \bibinfo{journal}{IEEE Transactions on Image Processing}
  \DOIprefix\doi{10.1109/83.902291}.
%Type = Article
\bibitem[{Chen et~al.(2010)Chen, van Eeden and Zidek}]{CHEN20101764}
\bibinfo{author}{Chen, J.}, \bibinfo{author}{van Eeden, C.},
  \bibinfo{author}{Zidek, J.}, \bibinfo{year}{2010}.
\newblock \bibinfo{title}{{Uncertainty and the conditional variance}}.
\newblock \bibinfo{journal}{Statistics and Probability Letters}
  \bibinfo{volume}{80}, \bibinfo{pages}{1764--1770}.
\newblock \URLprefix
  \url{https://www.sciencedirect.com/science/article/pii/S0167715210002154},
  \DOIprefix\doi{https://doi.org/10.1016/j.spl.2010.07.021}.
%Type = Inproceedings
\bibitem[{Chen et~al.(2020)Chen, Lian, Jiao, Wang, Gao and
  Lingling}]{chen2020supervised}
\bibinfo{author}{Chen, X.}, \bibinfo{author}{Lian, Y.}, \bibinfo{author}{Jiao,
  L.}, \bibinfo{author}{Wang, H.}, \bibinfo{author}{Gao, Y.},
  \bibinfo{author}{Lingling, S.}, \bibinfo{year}{2020}.
\newblock \bibinfo{title}{Supervised edge attention network for accurate image
  instance segmentation}, in: \bibinfo{booktitle}{European Conference on
  Computer Vision}, \bibinfo{organization}{Springer}. pp.
  \bibinfo{pages}{617--631}.
\newblock \DOIprefix\doi{https://doi.org/10.1007/978-3-030-58583-9-37}.
%Type = Article
\bibitem[{Cheng and Zhou(2021)}]{Cheng_2021}
\bibinfo{author}{Cheng, W.}, \bibinfo{author}{Zhou, Y.}, \bibinfo{year}{2021}.
\newblock \bibinfo{title}{Automatic pavement crack detection based on
  hierarchical feature augmentation}.
\newblock \bibinfo{journal}{International Conference on Artificial Intelligence
  and Information Systems} \DOIprefix\doi{10.1145/3469213.3470392}.
%Type = Article
\bibitem[{Deng et~al.(2021)Deng, Lu and Lee}]{deng2021imaging}
\bibinfo{author}{Deng, J.}, \bibinfo{author}{Lu, Y.}, \bibinfo{author}{Lee,
  V.C.S.}, \bibinfo{year}{2021}.
\newblock \bibinfo{title}{Imaging-based crack detection on concrete surfaces
  using you only look once network}.
\newblock \bibinfo{journal}{Structural Health Monitoring} \bibinfo{volume}{20},
  \bibinfo{pages}{484--499}.
\newblock \DOIprefix\doi{https://doi.org/10.1177/1475921720938486}.
%Type = Article
\bibitem[{Fan et~al.()Fan, Zhang and Yu}]{FAN2023109088}
\bibinfo{author}{Fan, X.}, \bibinfo{author}{Zhang, X.}, \bibinfo{author}{Yu,
  X.B.}, .
\newblock \bibinfo{title}{Uncertainty quantification of a deep learning model
  for failure rate prediction of water distribution networks}.
\newblock \bibinfo{journal}{Reliability Engineering and System Safety}
  \DOIprefix\doi{https://doi.org/10.1016/j.ress.2023.109088}.
%Type = Article
\bibitem[{Fan et~al.(2018)Fan, Wu, Lu and Li}]{fan2018automatic}
\bibinfo{author}{Fan, Z.}, \bibinfo{author}{Wu, Y.}, \bibinfo{author}{Lu, J.},
  \bibinfo{author}{Li, W.}, \bibinfo{year}{2018}.
\newblock \bibinfo{title}{Automatic pavement crack detection based on
  structured prediction with the convolutional neural network}.
\newblock \bibinfo{journal}{arXiv preprint} \DOIprefix\doi{arXiv:1802.02208}.
%Type = Inproceedings
\bibitem[{Gal and Ghahramani(2016)}]{gal2016dropout}
\bibinfo{author}{Gal, Y.}, \bibinfo{author}{Ghahramani, Z.},
  \bibinfo{year}{2016}.
\newblock \bibinfo{title}{Dropout as a bayesian approximation: Representing
  model uncertainty in deep learning}, in: \bibinfo{booktitle}{international
  conference on machine learning}, \bibinfo{organization}{PMLR}. pp.
  \bibinfo{pages}{1050--1059}.
\newblock \URLprefix \url{https://proceedings.mlr.press/v48/gal16.html},
  \DOIprefix\doi{arXiv:1506.02142}.
%Type = Inproceedings
\bibitem[{Gal et~al.(2017)Gal, Hron and Kendall}]{YarinConcrete2017}
\bibinfo{author}{Gal, Y.}, \bibinfo{author}{Hron, J.},
  \bibinfo{author}{Kendall, A.}, \bibinfo{year}{2017}.
\newblock \bibinfo{title}{Concrete dropout}, in: \bibinfo{editor}{Guyon, I.},
  \bibinfo{editor}{Luxburg, U.V.}, \bibinfo{editor}{Bengio, S.},
  \bibinfo{editor}{Wallach, H.}, \bibinfo{editor}{Fergus, R.},
  \bibinfo{editor}{Vishwanathan, S.}, \bibinfo{editor}{Garnett, R.} (Eds.),
  \bibinfo{booktitle}{Advances in Neural Information Processing Systems},
  \bibinfo{publisher}{Curran Associates, Inc.}. p.~\bibinfo{pages}{0}.
\newblock \DOIprefix\doi{https://doi.org/10.48550/arXiv.1705.07832}.
%Type = Article
\bibitem[{Girshick(2015)}]{Girshick_2015}
\bibinfo{author}{Girshick, R.}, \bibinfo{year}{2015}.
\newblock \bibinfo{title}{Fast r-cnn}.
\newblock \bibinfo{journal}{IEEE International Conference on Computer Vision}
  \DOIprefix\doi{10.1109/iccv.2015.169}.
%Type = Misc
\bibitem[{Groenendijk et~al.(2020)Groenendijk, Karaoglu, Gevers and
  Mensink}]{groenendijk2020multiloss}
\bibinfo{author}{Groenendijk, R.}, \bibinfo{author}{Karaoglu, S.},
  \bibinfo{author}{Gevers, T.}, \bibinfo{author}{Mensink, T.},
  \bibinfo{year}{2020}.
\newblock \bibinfo{title}{Multi-loss weighting with coefficient of variations}.
\newblock \DOIprefix\doi{https://doi.org/10.48550/arXiv.2009.01717},
  \href{http://arxiv.org/abs/2009.01717}{{\tt arXiv:2009.01717}}.
%Type = Misc
\bibitem[{Guo et~al.(2017)Guo, Pleiss, Sun and
  Weinberger}]{https://doi.org/10.48550/arxiv.1706.04599}
\bibinfo{author}{Guo, C.}, \bibinfo{author}{Pleiss, G.}, \bibinfo{author}{Sun,
  Y.}, \bibinfo{author}{Weinberger, K.Q.}, \bibinfo{year}{2017}.
\newblock \bibinfo{title}{On calibration of modern neural networks}.
\newblock \URLprefix \url{https://arxiv.org/abs/1706.04599},
  \DOIprefix\doi{10.48550/ARXIV.1706.04599}.
%Type = Article
\bibitem[{Guo et~al.(2021)Guo, Markoni and Lee}]{guo2021barnet}
\bibinfo{author}{Guo, J.M.}, \bibinfo{author}{Markoni, H.},
  \bibinfo{author}{Lee, J.D.}, \bibinfo{year}{2021}.
\newblock \bibinfo{title}{Barnet: Boundary aware refinement network for crack
  detection}.
\newblock \bibinfo{journal}{IEEE Transactions on Intelligent Transportation
  Systems} \DOIprefix\doi{10.1109/TITS.2021.3069135}.
%Type = Article
\bibitem[{Hacıefendioğlu and Başağa(2021)}]{Hacıefendioğlu_2021}
\bibinfo{author}{Hacıefendioğlu, K.}, \bibinfo{author}{Başağa, H.B.},
  \bibinfo{year}{2021}.
\newblock \bibinfo{title}{Concrete road crack detection using deep
  learning-based faster r-cnn method}.
\newblock \bibinfo{journal}{Iranian Journal of Science and
  Technology-Transactions of Civil Engineering}
  \DOIprefix\doi{10.1007/s40996-021-00671-2}.
%Type = Article
\bibitem[{He et~al.(2016)He, Zhang, Ren and Sun}]{He_2016}
\bibinfo{author}{He, K.}, \bibinfo{author}{Zhang, X.}, \bibinfo{author}{Ren,
  S.}, \bibinfo{author}{Sun, J.}, \bibinfo{year}{2016}.
\newblock \bibinfo{title}{Deep residual learning for image recognition}.
\newblock \bibinfo{journal}{Computer Vision and Pattern Recognition}
  \DOIprefix\doi{10.1109/cvpr.2016.90}.
%Type = Article
\bibitem[{Kato et~al.(2022)Kato, Hino, Kume and Nobuhara}]{Kato_2022}
\bibinfo{author}{Kato, S.}, \bibinfo{author}{Hino, T.}, \bibinfo{author}{Kume,
  S.}, \bibinfo{author}{Nobuhara, H.}, \bibinfo{year}{2022}.
\newblock \bibinfo{title}{Crack detection from weld bend test images using
  r-cnn}.
\newblock \bibinfo{journal}{International Conference on P2P, Parallel, Grid,
  Cloud and Internet Computing}
  \DOIprefix\doi{https://doi.org/10.1007/978-3-030-89899-1}.
%Type = Article
\bibitem[{Kendall and Gal(2017)}]{kendall2017uncertainties}
\bibinfo{author}{Kendall, A.}, \bibinfo{author}{Gal, Y.}, \bibinfo{year}{2017}.
\newblock \bibinfo{title}{What uncertainties do we need in bayesian deep
  learning for computer vision?}
\newblock \bibinfo{journal}{CoRR} \bibinfo{volume}{abs/1703.04977}.
\newblock \DOIprefix\doi{arXiv:1703.04977}.
%Type = Article
\bibitem[{Kirschke et~al.(1992)Kirschke, Velinsky and Velinsky}]{Kirschke_1992}
\bibinfo{author}{Kirschke, K.R.}, \bibinfo{author}{Velinsky, S.A.},
  \bibinfo{author}{Velinsky, S.A.}, \bibinfo{year}{1992}.
\newblock \bibinfo{title}{Histogram‐based approach for automated
  pavement‐crack sensing}.
\newblock \bibinfo{journal}{Journal of Transportation Engineering-asce}
  \DOIprefix\doi{10.1061/(asce)0733-947x(1992)118:5(700)}.
%Type = Article
\bibitem[{Kr{\"a}henb{\"u}hl and Koltun(2011)}]{krahenbuhl2011efficient}
\bibinfo{author}{Kr{\"a}henb{\"u}hl, P.}, \bibinfo{author}{Koltun, V.},
  \bibinfo{year}{2011}.
\newblock \bibinfo{title}{Efficient inference in fully connected crfs with
  gaussian edge potentials}.
\newblock \bibinfo{journal}{Advances in neural information processing systems}
  \bibinfo{volume}{24}.
\newblock \DOIprefix\doi{https://doi.org/10.48550/arXiv.1210.5644}.
%Type = Article
\bibitem[{Lee et~al.(2022)Lee, Yoon, Park, Eum and Cho}]{lee2022demonstration}
\bibinfo{author}{Lee, D.}, \bibinfo{author}{Yoon, S.}, \bibinfo{author}{Park,
  J.}, \bibinfo{author}{Eum, S.}, \bibinfo{author}{Cho, H.},
  \bibinfo{year}{2022}.
\newblock \bibinfo{title}{Demonstration of model-assisted probability of
  detection framework for ultrasonic inspection of cracks in compressor
  blades}.
\newblock \bibinfo{journal}{NDT \& E International} \bibinfo{volume}{128},
  \bibinfo{pages}{102618}.
\newblock \DOIprefix\doi{https://doi.org/10.1016/j.ndteint.2022.10261}.
%Type = Article
\bibitem[{Lee et~al.(2017)Lee, Jun, Cho, Lee, Kim, Seo and Kim}]{lee2017deep}
\bibinfo{author}{Lee, J.G.}, \bibinfo{author}{Jun, S.}, \bibinfo{author}{Cho,
  Y.W.}, \bibinfo{author}{Lee, H.}, \bibinfo{author}{Kim, G.B.},
  \bibinfo{author}{Seo, J.B.}, \bibinfo{author}{Kim, N.}, \bibinfo{year}{2017}.
\newblock \bibinfo{title}{Deep learning in medical imaging: general overview}.
\newblock \bibinfo{journal}{Korean journal of radiology} \bibinfo{volume}{18},
  \bibinfo{pages}{570--584}.
\newblock \DOIprefix\doi{10.3348/kjr.2017.18.4.570}.
%Type = Misc
\bibitem[{Liu et~al.(2020)Liu, Lin, Padhy, Tran, Bedrax-Weiss and
  Lakshminarayanan}]{lakshminarayanan2020simple}
\bibinfo{author}{Liu, J.Z.}, \bibinfo{author}{Lin, Z.}, \bibinfo{author}{Padhy,
  S.}, \bibinfo{author}{Tran, D.}, \bibinfo{author}{Bedrax-Weiss, T.},
  \bibinfo{author}{Lakshminarayanan, B.}, \bibinfo{year}{2020}.
\newblock \bibinfo{title}{Simple and principled uncertainty estimation with
  deterministic deep learning via distance awareness}.
\newblock \href{http://arxiv.org/abs/2006.10108}{{\tt arXiv:2006.10108}}.
%Type = Article
\bibitem[{Liu et~al.(2019)Liu, Yao, Lu, Xie and Li}]{LIU2019139}
\bibinfo{author}{Liu, Y.}, \bibinfo{author}{Yao, J.}, \bibinfo{author}{Lu, X.},
  \bibinfo{author}{Xie, R.}, \bibinfo{author}{Li, L.}, \bibinfo{year}{2019}.
\newblock \bibinfo{title}{Deepcrack: A deep hierarchical feature learning
  architecture for crack segmentation}.
\newblock \bibinfo{journal}{Neurocomputing} \bibinfo{volume}{338},
  \bibinfo{pages}{139--153}.
\newblock \URLprefix
  \url{https://www.sciencedirect.com/science/article/pii/S0925231219300566},
  \DOIprefix\doi{https://doi.org/10.1016/j.neucom.2019.01.036}.
%Type = Article
\bibitem[{Long et~al.(2017)Long, Shelhamer and Darrell}]{long2015fully}
\bibinfo{author}{Long, J.}, \bibinfo{author}{Shelhamer, E.},
  \bibinfo{author}{Darrell, T.}, \bibinfo{year}{2017}.
\newblock \bibinfo{title}{Fully convolutional networks for semantic
  segmentation}.
\newblock \bibinfo{journal}{IEEE Transactions on Pattern Analysis and Machine
  Intelligence} \DOIprefix\doi{10.1109/TPAMI.2016.2572683}.
%Type = Article
\bibitem[{Mao et~al.(2020)Mao, Chen, Ping, Ping, Ping and Hao}]{Mao_2020}
\bibinfo{author}{Mao, Y.}, \bibinfo{author}{Chen, J.}, \bibinfo{author}{Ping,
  P.}, \bibinfo{author}{Ping, P.}, \bibinfo{author}{Ping, P.},
  \bibinfo{author}{Hao, C.}, \bibinfo{year}{2020}.
\newblock \bibinfo{title}{Crack detection with multi-task enhanced faster r-cnn
  model}.
\newblock \bibinfo{journal}{International Conference on Big Data Computing
  Service and Applications}
  \DOIprefix\doi{10.1109/bigdataservice49289.2020.00038}.
%Type = Article
\bibitem[{Mao-de et~al.(2007)Mao-de, Shaobo, Kun and Yuyao}]{Mao-de_2007}
\bibinfo{author}{Mao-de, Y.}, \bibinfo{author}{Shaobo, B.},
  \bibinfo{author}{Kun, X.}, \bibinfo{author}{Yuyao, H.}, \bibinfo{year}{2007}.
\newblock \bibinfo{title}{Pavement crack detection and analysis for high-grade
  highway}.
\newblock \bibinfo{journal}{International Conference on Electronic Measurement
  and Instruments} \DOIprefix\doi{10.1109/icemi.2007.4351202}.
%Type = Article
\bibitem[{McFarland and DeCarlo(2020)}]{McFarland_2020}
\bibinfo{author}{McFarland, J.}, \bibinfo{author}{DeCarlo, E.C.},
  \bibinfo{year}{2020}.
\newblock \bibinfo{title}{A monte carlo framework for probabilistic analysis
  and variance decomposition with distribution parameter uncertainty}.
\newblock \bibinfo{journal}{Reliability Engineering and System Safety}
  \DOIprefix\doi{10.1016/j.ress.2020.106807}.
%Type = Article
\bibitem[{Moradi et~al.(2022)Moradi, Cofre-Martel, Droguett, Modarres and
  Groth}]{Moradi_2022}
\bibinfo{author}{Moradi, R.}, \bibinfo{author}{Cofre-Martel, S.},
  \bibinfo{author}{Droguett, E.L.}, \bibinfo{author}{Modarres, M.},
  \bibinfo{author}{Groth, K.M.}, \bibinfo{year}{2022}.
\newblock \bibinfo{title}{Integration of deep learning and bayesian networks
  for condition and operation risk monitoring of complex engineering systems}.
\newblock \bibinfo{journal}{Reliability Engineering {\&} System Safety}
  \DOIprefix\doi{10.1016/j.ress.2022.108433}.
%Type = Inproceedings
\bibitem[{Naeini et~al.(2015)Naeini, Cooper and
  Hauskrecht}]{naeini2015obtaining}
\bibinfo{author}{Naeini, M.P.}, \bibinfo{author}{Cooper, G.F.},
  \bibinfo{author}{Hauskrecht, M.}, \bibinfo{year}{2015}.
\newblock \bibinfo{title}{Obtaining well calibrated probabilities using
  bayesian binning}, in: \bibinfo{booktitle}{Proceedings of the Twenty-Ninth
  AAAI Conference on Artificial Intelligence}, \bibinfo{publisher}{AAAI Press}.
  p. \bibinfo{pages}{2901–2907}.
\newblock \DOIprefix\doi{https://doi.org/10.1609/aaai.v29i1.9602}.
%Type = Article
\bibitem[{Pang et~al.(2022)Pang, Zhao, Hu, Yan and Liu}]{pang2022bayesian}
\bibinfo{author}{Pang, Y.}, \bibinfo{author}{Zhao, X.}, \bibinfo{author}{Hu,
  J.}, \bibinfo{author}{Yan, H.}, \bibinfo{author}{Liu, Y.},
  \bibinfo{year}{2022}.
\newblock \bibinfo{title}{Bayesian spatio-temporal graph transformer network
  (b-star) for multi-aircraft trajectory prediction}.
\newblock \bibinfo{journal}{Knowledge-Based Systems} ,
  \bibinfo{pages}{108998}\DOIprefix\doi{https://doi.org/10.1016/j.knosys.2022.108998}.
%Type = Article
\bibitem[{Pang et~al.(2021)Pang, Zhao, Yan and Liu}]{pang2021data}
\bibinfo{author}{Pang, Y.}, \bibinfo{author}{Zhao, X.}, \bibinfo{author}{Yan,
  H.}, \bibinfo{author}{Liu, Y.}, \bibinfo{year}{2021}.
\newblock \bibinfo{title}{Data-driven trajectory prediction with weather
  uncertainties: A bayesian deep learning approach}.
\newblock \bibinfo{journal}{Transportation Research Part C: Emerging
  Technologies} \bibinfo{volume}{130}, \bibinfo{pages}{103326}.
\newblock \DOIprefix\doi{https://doi.org/10.1016/j.trc.2021.103326}.
%Type = Article
\bibitem[{Pyle et~al.(2022)Pyle, Hughes, Ali and Wilcox}]{Pyle_null}
\bibinfo{author}{Pyle, R.J.}, \bibinfo{author}{Hughes, R.R.},
  \bibinfo{author}{Ali, A.A.S.}, \bibinfo{author}{Wilcox, P.D.},
  \bibinfo{year}{2022}.
\newblock \bibinfo{title}{Uncertainty quantification for deep learning in
  ultrasonic crack characterization.}
\newblock \bibinfo{journal}{IEEE Transactions on Ultrasonics Ferroelectrics and
  Frequency Control} \DOIprefix\doi{10.1109/tuffc.2022.3176926}.
%Type = Article
\bibitem[{Redmon et~al.(2016)Redmon, Divvala, Girshick and
  Farhadi}]{Redmon_2016}
\bibinfo{author}{Redmon, J.}, \bibinfo{author}{Divvala, S.K.},
  \bibinfo{author}{Girshick, R.}, \bibinfo{author}{Farhadi, A.},
  \bibinfo{year}{2016}.
\newblock \bibinfo{title}{You only look once: Unified, real-time object
  detection}.
\newblock \bibinfo{journal}{Computer Vision and Pattern Recognition}
  \DOIprefix\doi{10.1109/cvpr.2016.91}.
%Type = Article
\bibitem[{Ren et~al.(2015)Ren, He, Girshick and Sun}]{Ren_2015}
\bibinfo{author}{Ren, S.}, \bibinfo{author}{He, K.}, \bibinfo{author}{Girshick,
  R.}, \bibinfo{author}{Sun, J.}, \bibinfo{year}{2015}.
\newblock \bibinfo{title}{Faster r-cnn: Towards real-time object detection with
  region proposal networks}.
\newblock \bibinfo{journal}{arXiv: Computer Vision and Pattern Recognition}
  \DOIprefix\doi{10.1109/tpami.2016.2577031}.
%Type = Article
\bibitem[{Ronneberger et~al.(2015)Ronneberger, Fischer and
  Brox}]{Ronneberger_2015}
\bibinfo{author}{Ronneberger, O.}, \bibinfo{author}{Fischer, P.},
  \bibinfo{author}{Brox, T.}, \bibinfo{year}{2015}.
\newblock \bibinfo{title}{U-net: Convolutional networks for biomedical image
  segmentation}.
\newblock \bibinfo{journal}{arXiv: Computer Vision and Pattern Recognition}
  \DOIprefix\doi{10.1007/978-3-319-24574-4-28}.
%Type = Article
\bibitem[{Sajedi and Liang(2020)}]{Sajedi_2020}
\bibinfo{author}{Sajedi, S.O.}, \bibinfo{author}{Liang, X.},
  \bibinfo{year}{2020}.
\newblock \bibinfo{title}{Uncertainty‐assisted deep vision structural health
  monitoring}.
\newblock \bibinfo{journal}{Computer-aided Civil and Infrastructure
  Engineering} \DOIprefix\doi{10.1111/mice.12580}.
%Type = Article
\bibitem[{Seites-Rundlett et~al.(2021)Seites-Rundlett, Bashar, Torres-Machi and
  Corotis}]{Seites-Rundlett_2021}
\bibinfo{author}{Seites-Rundlett, W.}, \bibinfo{author}{Bashar, M.Z.},
  \bibinfo{author}{Torres-Machi, C.}, \bibinfo{author}{Corotis, R.B.},
  \bibinfo{year}{2021}.
\newblock \bibinfo{title}{Combined evidence model to enhance pavement condition
  prediction from highly uncertain sensor data}.
\newblock \bibinfo{journal}{Reliability Engineering {\&} System Safety}
  \DOIprefix\doi{10.1016/j.ress.2021.108031}.
%Type = Article
\bibitem[{Shi et~al.(2016)Shi, Cui, Qi, Meng and Chen}]{Shi2016}
\bibinfo{author}{Shi, Y.}, \bibinfo{author}{Cui, L.}, \bibinfo{author}{Qi, Z.},
  \bibinfo{author}{Meng, F.}, \bibinfo{author}{Chen, Z.}, \bibinfo{year}{2016}.
\newblock \bibinfo{title}{{Automatic road crack detection using random
  structured forests}}.
\newblock \bibinfo{journal}{IEEE Transactions on Intelligent Transportation
  Systems} \DOIprefix\doi{10.1109/TITS.2016.2552248}.
%Type = Article
\bibitem[{Srivastava et~al.(2014)Srivastava, Hinton, Krizhevsky, Sutskever and
  Salakhutdinov}]{srivastava2014dropout}
\bibinfo{author}{Srivastava, N.}, \bibinfo{author}{Hinton, G.},
  \bibinfo{author}{Krizhevsky, A.}, \bibinfo{author}{Sutskever, I.},
  \bibinfo{author}{Salakhutdinov, R.}, \bibinfo{year}{2014}.
\newblock \bibinfo{title}{Dropout: a simple way to prevent neural networks from
  overfitting}.
\newblock \bibinfo{journal}{The journal of machine learning research}
  \bibinfo{volume}{15}, \bibinfo{pages}{1929--1958}.
\newblock \URLprefix \url{http://jmlr.org/papers/v15/srivastava14a.html},
  \DOIprefix\doi{10.5555/2627435.2670313}.
%Type = Article
\bibitem[{Sun et~al.(2023)Sun, Zhu, Qiu, Liu, Xiang and
  Xuan}]{sun2023nonlinear}
\bibinfo{author}{Sun, D.}, \bibinfo{author}{Zhu, W.}, \bibinfo{author}{Qiu,
  X.}, \bibinfo{author}{Liu, L.}, \bibinfo{author}{Xiang, Y.},
  \bibinfo{author}{Xuan, F.Z.}, \bibinfo{year}{2023}.
\newblock \bibinfo{title}{Nonlinear ultrasonic detection of closed cracks in
  metal plates with phase-velocity mismatching}.
\newblock \bibinfo{journal}{NDT \& E International} ,
  \bibinfo{pages}{102788}\DOIprefix\doi{https://doi.org/10.1016/j.ndteint.2023.102788}.
%Type = Article
\bibitem[{Te~Han(2022)}]{HAN2022108648}
\bibinfo{author}{Te~Han, Y.F.L.}, \bibinfo{year}{2022}.
\newblock \bibinfo{title}{Out-of-distribution detection-assisted trustworthy
  machinery fault diagnosis approach with uncertainty-aware deep ensembles}.
\newblock \bibinfo{journal}{Reliability Engineering and System Safety}
  \DOIprefix\doi{https://doi.org/10.1016/j.ress.2022.108648}.
%Type = Article
\bibitem[{Tohme et~al.(2022)Tohme, Vanslette and Youcef-Toumi}]{Tohme_2022}
\bibinfo{author}{Tohme, T.}, \bibinfo{author}{Vanslette, K.},
  \bibinfo{author}{Youcef-Toumi, K.}, \bibinfo{year}{2022}.
\newblock \bibinfo{title}{Reliable neural networks for regression uncertainty
  estimation}.
\newblock \bibinfo{journal}{Reliability Engineering and System Safety}
  \DOIprefix\doi{10.1016/j.ress.2022.108811}.
%Type = Misc
\bibitem[{Verdoja and Kyrki(2021)}]{verdoja2021notes}
\bibinfo{author}{Verdoja, F.}, \bibinfo{author}{Kyrki, V.},
  \bibinfo{year}{2021}.
\newblock \bibinfo{title}{Notes on the behavior of mc dropout}.
\newblock \DOIprefix\doi{https://doi.org/10.48550/arXiv.2008.02627},
  \href{http://arxiv.org/abs/2008.02627}{{\tt arXiv:2008.02627}}.
%Type = Article
\bibitem[{Wang et~al.(2021)Wang, Zhang, Cui, Liu, Ren, Yang, Xie, Hua, Bao and
  Xu}]{DBLP:journals/corr/abs-2102-02696}
\bibinfo{author}{Wang, C.}, \bibinfo{author}{Zhang, Y.}, \bibinfo{author}{Cui,
  M.}, \bibinfo{author}{Liu, J.}, \bibinfo{author}{Ren, P.},
  \bibinfo{author}{Yang, Y.}, \bibinfo{author}{Xie, X.}, \bibinfo{author}{Hua,
  X.}, \bibinfo{author}{Bao, H.}, \bibinfo{author}{Xu, W.},
  \bibinfo{year}{2021}.
\newblock \bibinfo{title}{Active boundary loss for semantic segmentation}.
\newblock \bibinfo{journal}{CoRR} \bibinfo{volume}{abs/2102.02696}.
\newblock \URLprefix \url{https://arxiv.org/abs/2102.02696},
  \DOIprefix\doi{https://doi.org/10.48550/arXiv.2102.02696},
  \href{http://arxiv.org/abs/2102.02696}{{\tt arXiv:2102.02696}}.
%Type = Article
\bibitem[{Wang et~al.(2018)Wang, Hwang, Rose and Wallace}]{wang2018uncertainty}
\bibinfo{author}{Wang, G.}, \bibinfo{author}{Hwang, J.N.},
  \bibinfo{author}{Rose, C.}, \bibinfo{author}{Wallace, F.},
  \bibinfo{year}{2018}.
\newblock \bibinfo{title}{Uncertainty-based active learning via sparse modeling
  for image classification}.
\newblock \bibinfo{journal}{IEEE Transactions on Image Processing}
  \bibinfo{volume}{28}, \bibinfo{pages}{316--329}.
%Type = Article
\bibitem[{Woods and Allen(1989)}]{Woods_1989}
\bibinfo{author}{Woods, P.W.}, \bibinfo{author}{Allen, P.D.},
  \bibinfo{year}{1989}.
\newblock \bibinfo{title}{A cue generator for crack detection}.
\newblock \bibinfo{journal}{Image and Vision Computing}
  \DOIprefix\doi{10.1016/0262-8856(89)90030-9}.
%Type = Article
\bibitem[{Yang et~al.(2020)Yang, Yang, Yang, Zhang, Zhang, Yu, Prokhorov, Mei
  and Ling}]{Yang_2020}
\bibinfo{author}{Yang, F.}, \bibinfo{author}{Yang, F.}, \bibinfo{author}{Yang,
  F.}, \bibinfo{author}{Zhang, L.}, \bibinfo{author}{Zhang, L.},
  \bibinfo{author}{Yu, S.}, \bibinfo{author}{Prokhorov, D.V.},
  \bibinfo{author}{Mei, X.}, \bibinfo{author}{Ling, H.}, \bibinfo{year}{2020}.
\newblock \bibinfo{title}{Feature pyramid and hierarchical boosting network for
  pavement crack detection}.
\newblock \bibinfo{journal}{IEEE Transactions on Intelligent Transportation
  Systems} \DOIprefix\doi{10.1109/tits.2019.2910595}.
%Type = Article
\bibitem[{Yang et~al.(2019)Yang, Zhang, Yu, Prokhorov, Mei and
  Ling}]{yang2019feature}
\bibinfo{author}{Yang, F.}, \bibinfo{author}{Zhang, L.}, \bibinfo{author}{Yu,
  S.}, \bibinfo{author}{Prokhorov, D.}, \bibinfo{author}{Mei, X.},
  \bibinfo{author}{Ling, H.}, \bibinfo{year}{2019}.
\newblock \bibinfo{title}{Feature pyramid and hierarchical boosting network for
  pavement crack detection}.
\newblock \bibinfo{journal}{IEEE Transactions on Intelligent Transportation
  Systems} \bibinfo{volume}{21}, \bibinfo{pages}{1525--1535}.
\newblock \DOIprefix\doi{10.1109/TITS.2019.2910595}.
%Type = Inproceedings
\bibitem[{Yang and Loog(2016)}]{yang2016active}
\bibinfo{author}{Yang, Y.}, \bibinfo{author}{Loog, M.}, \bibinfo{year}{2016}.
\newblock \bibinfo{title}{Active learning using uncertainty information}, in:
  \bibinfo{booktitle}{2016 23rd International Conference on Pattern Recognition
  (ICPR)}, \bibinfo{organization}{IEEE}. pp. \bibinfo{pages}{2646--2651}.
%Type = Article
\bibitem[{Zhang et~al.(2018)Zhang, B{\"u}tepage, Kjellstr{\"o}m and
  Mandt}]{zhang2018advances}
\bibinfo{author}{Zhang, C.}, \bibinfo{author}{B{\"u}tepage, J.},
  \bibinfo{author}{Kjellstr{\"o}m, H.}, \bibinfo{author}{Mandt, S.},
  \bibinfo{year}{2018}.
\newblock \bibinfo{title}{Advances in variational inference}.
\newblock \bibinfo{journal}{IEEE transactions on pattern analysis and machine
  intelligence} \bibinfo{volume}{41}, \bibinfo{pages}{2008--2026}.
\newblock \DOIprefix\doi{10.1109/TPAMI.2018.2889774}.
%Type = Article
\bibitem[{Zhou et~al.(2022)Zhou, Han and Droguett}]{Zhou_2022}
\bibinfo{author}{Zhou, T.}, \bibinfo{author}{Han, T.},
  \bibinfo{author}{Droguett, E.L.}, \bibinfo{year}{2022}.
\newblock \bibinfo{title}{Towards trustworthy machine fault diagnosis: A
  probabilistic bayesian deep learning framework}.
\newblock \bibinfo{journal}{Reliability Engineering {\&} System Safety}
  \DOIprefix\doi{10.1016/j.ress.2022.108525}.
%Type = Article
\bibitem[{Zhu et~al.(2022)Zhu, Chen, Peng and Ye}]{Zhu_2022}
\bibinfo{author}{Zhu, R.}, \bibinfo{author}{Chen, Y.}, \bibinfo{author}{Peng,
  W.}, \bibinfo{author}{Ye, Z.S.}, \bibinfo{year}{2022}.
\newblock \bibinfo{title}{Bayesian deep-learning for rul prediction: An active
  learning perspective}.
\newblock \bibinfo{journal}{Reliability Engineering {\&} System Safety}
  \DOIprefix\doi{10.1016/j.ress.2022.108758}.
%Type = Article
\bibitem[{Zou et~al.(2018)Zou, Zhang, Li, Qi, Wang and Wang}]{zou2018deepcrack}
\bibinfo{author}{Zou, Q.}, \bibinfo{author}{Zhang, Z.}, \bibinfo{author}{Li,
  Q.}, \bibinfo{author}{Qi, X.}, \bibinfo{author}{Wang, Q.},
  \bibinfo{author}{Wang, S.}, \bibinfo{year}{2018}.
\newblock \bibinfo{title}{Deepcrack: Learning hierarchical convolutional
  features for crack detection}.
\newblock \bibinfo{journal}{IEEE Transactions on Image Processing}
  \bibinfo{volume}{28}, \bibinfo{pages}{1498--1512}.
\newblock \DOIprefix\doi{10.1109/TIP.2018.2878966}.

\end{thebibliography}


\end{document}